\documentclass[sigconf, table, prologe]{acmart}
\usepackage{listings}
\usepackage{booktabs}

\usepackage{array}
\usepackage{multirow}
\usepackage{enumitem}
\usepackage[linesnumbered,ruled,vlined]{algorithm2e}
\usepackage{lineno}
\usepackage[toc,page]{appendix}
\DeclareUnicodeCharacter{2606}{\ensuremath{\star}}
\usepackage[utf8]{inputenc}
\usepackage[T1]{fontenc}
\usepackage{tabularx}
\newcommand{\myrowcolour}{\rowcolor[gray]{0.95}}
\usepackage{makecell}

\renewcommand\footnotetextcopyrightpermission[1]{} 

\AtBeginDocument{%
  \providecommand\BibTeX{{%
    \normalfont B\kern-0.5em{\scshape i\kern-0.25em b}\kern-0.8em\TeX}}}

\graphicspath{{./images/}} 
\begin{document}


\title{Multi-Agent Reinforcement Learning for Dynamic Pricing in Supply Chains: Benchmarking Strategic Agent Behaviours under Realistically Simulated Market Conditions}

\author{
  Thomas Hazenberg$^{1}$, 
  Yao Ma$^{1}$*, 
  Seyed Sahand Mohammadi Ziabari$^{1,2}$,
  and Marijn van Rijswijk$^{3}$
}

\affiliation{%
  \institution{
  $^{1}$Informatics Institute, University of Amsterdam, 1098XH Science Park, Amsterdam, The Netherlands}
  \country{}
}

\affiliation{%
  \institution{$^{2}$Department of Computer Science and Technology, SUNY Empire State University, Saratoga Springs, NY, USA}
   \country{}
}

\affiliation{%
  \institution{
  $^{3}$KPMG N.V, 1186 DS Amstelveen, The Netherlands}
  \country{}
}

\email{thomas.hazenberg@student.uva.nl, y.ma3@uva.nl,sahand.ziabari@sunyempire.edu, vanrijswijk.marijn@kpmg.nl}




\pagestyle{plain}

\setcounter{page}{0}

\begin{abstract}
This study investigates how Multi-Agent Reinforcement Learning (MARL) can improve dynamic pricing strategies in supply chains, particularly in contexts where traditional ERP systems rely on static, rule-based approaches that overlook strategic interactions among market actors. While recent research has applied reinforcement learning to pricing, most implementations remain single-agent and fail to model the interdependent nature of real-world supply chains. This study addresses that gap by evaluating the performance of three MARL algorithms: MADDPG, MADQN, and QMIX against static rule-based baselines, within a simulated environment informed by real e-commerce transaction data and a LightGBM demand prediction model. Results show that rule-based agents achieve near-perfect fairness (Jain’s Index: \textit{0.9896}) and the highest price stability (volatility: \textit{0.024}), but they fully lack competitive dynamics. Among MARL agents, MADQN exhibits the most aggressive pricing behaviour, with the highest volatility and the lowest fairness (\textit{0.5844}). MADDPG provides a more balanced approach, supporting market competition (share volatility: \textit{9.5 pp}) while maintaining relatively high fairness (\textit{0.8819}) and stable pricing. These findings suggest that MARL introduces emergent strategic behaviour not captured by static pricing rules and may inform future developments in dynamic pricing.

\end{abstract}

\maketitle

\section{Introduction}
\label{sec:introduction}

As customers demand more advanced functionalities from enterprise resource planning (ERP) systems and their vendors to manage inventories, manufacturers, suppliers, distributors, and retailers, the need for an intelligent dynamic pricing mechanism becomes more critical \cite{tarn2002exploring, emma2024enterprise, smith2024dynamic}. ERP systems efficiently integrate and automate core supply chain functions such as procurement, logistics, and financial planning, but their capacity to optimize dynamic pricing stays underutilized. Modern supply chains are highly interconnected and complex, with multiple stakeholders each making pricing and inventory decisions \cite{jain2008managing}. Most of the entities in these chains still rely on static pricing models. These are based on fixed rules, historical data, or cost-plus formulas, which fail to reflect real-time changes in demand, supply, or competitor behaviour \cite{hajji2012dynamic, li2015pricing}. As a result, these static pricing strategies could lead to revenue losses and missed market opportunities. For instance, Cachon and Feldman (2010) show that static or naïve dynamic pricing strategies can underperform by up to 22.6\% in revenue compared to dynamic pricing, with an average loss of 7\%, showing the real cost of failing adaptive pricing mechanisms \cite{cachon2010dynamic}. Unlocking this would allow businesses to set prices dynamically through demand, inventory, and other factors, thus enhancing profitability and competitiveness.

Dynamic pricing offers a data-driven alternative to static pricing by adjusting prices in response to market conditions. Integrating such strategies within ERP systems comes with several challenges. It requires seamless data exchange, real-time decision-making across actors, and scalable models. Current pricing solutions lack these capabilities, especially the ability to learn autonomously from evolving market dynamics \cite{menon2024erp}. Machine learning techniques have been increasingly used in ERP systems for tasks like forecasting, but there remains a lack of solutions that realistically model decentralized and interactive pricing decisions across multiple supply chain actors in a learning manner.

Recent research has explored Reinforcement Learning (RL) as a method to optimize pricing strategies \cite{wong2023deep}. Many RL-based models, such as Bayesian approaches, feature-based learning, and Q-learning, act in a single-agent setting, where pricing decisions are made isolated. This overlooks the interdependent nature of pricing decisions across supply chain entities \cite{jain2008managing, neto2005single}. When a retailer adjusts prices, it affects demand, thus influencing supplier production levels and inventory. Ignoring these dependencies could potentially lead to suboptimal pricing strategies.

To address this gap, we benchmark and evaluate different Multi-Agent Reinforcement Learning (MARL) algorithms for optimizing dynamic pricing in supply chains, built upon existing and extending MARL models. Unlike traditional RL methods, MARL involves multiple autonomous agents, where each represents a distinct entity with a product catalogue and pricing decisions to learn collaboratively or competitively. These agents then interact with a shared environment that is shaped by market conditions, competitor actions, and changes in supply and demand. They dynamically adjust their pricing strategies through reward-based learning, thereby potentially improving revenue generation and market adaptability.

In this context, MARL could be particularly suited for capturing the interdependent nature of pricing decisions in supply chains, which traditional single-agent RL methods fail to model \cite{busoniu2008comprehensive}. This study investigates the potential of MARL to enhance dynamic pricing strategies within ERP-integrated supply chains. Existing pricing mechanisms, particularly those based on static rules or heuristics, often lack the flexibility to adapt to volatile market environments and the strategic interactions between multiple pricing agents. By contrast, MARL offers a data-driven and adaptive alternative capable of capturing complex interdependencies and evolving agent behaviors in competitive and cooperative settings.

The research explores how various MARL algorithms, each with different learning dynamics and coordination mechanisms, can improve the adaptability, stability, and responsiveness of pricing. Comparative evaluation is conducted against traditional rule-based approaches to assess performance in environments characterized by fluctuating demand, customer segmentation, and supply-side variability.

In addition, the study examines how the efficacy of MARL-based strategies is influenced by underlying market dynamics, such as volatility, agent heterogeneity, and feedback latency. Through a series of controlled experiments, this work highlights the strengths and limitations of reinforcement learning in operational pricing tasks and contributes a practical framework for the deployment of MARL in enterprise-level decision support systems.

The remainder of this research is organized as follows. Section~\ref{sec:related_work} reviews relevant literature on pricing algorithms, reinforcement learning approaches to pricing, and challenges in applying MARL to supply chain contexts. Section~\ref{sec:methodology} details the methodological framework, including pre-processing of the data set, the demand model, the simulation environment, agent architectures, and the evaluation setup. Section~\ref{sec:results} presents the experimental results, comparing MARL strategies to rule-based baselines. Section~\ref{sec:discussion} interprets the findings in light of prior work, explores trade-offs, and discusses limitations. Finally, Section~\ref{sec:conclusion} concludes the study and outlines directions for future research.

\section{Related Work}
\label{sec:related_work}
Prior studies denote the importance of dynamic pricing in supply chain revenue optimization and market adaptability, but its implementation to ERP systems remains fairly limited \cite{kim2015dynamic, han2019multi}. Recent work has started exploring RL for pricing optimization, but it mostly focuses on single-agent configurations. This research aims to cross this gap by using MARL to enhance pricing strategies. To position this study in the existing literature, this section focuses on three areas: (1) Pricing Algorithms, covering both static and dynamic pricing; (2) Reinforcement Learning for Pricing Optimization, distinguishing between single-agent RL and MARL; and (3) Challenges in MARL, discussing computational complexity, scalability, and data limitations in applying MARL to pricing environments.

\subsection{Pricing Algorithms}
To build upon existing pricing methodologies, we explore static pricing, which relies on fixed or rule-based methods, and dynamic pricing, which uses machine learning to adjust prices. These pricing approaches are fundamental in supply chain management, providing a basis for more advanced RL-based pricing strategies.
 
\subsubsection{Static Pricing}
Traditional pricing relies on rule-based methods, cost-plus pricing, and historical trends. These models typically set fixed prices based on predefined formulas rather than adapting dynamically to market conditions. Rule-based pricing, often used in retail, adjusts prices by, for instance, undercutting the lowest competitor price or following markup rules. In spite of their simplicity, rule-based pricing remains widely used, particularly among Amazon, Walmart, and eBay \cite{wang2023algorithms, chen2021multiobjective}. Still, their lack of adaptability could lead to suboptimal revenue and potential algorithmic complicity in generally competitive markets. Another common approach, cost-plus pricing, determines prices by adding a fixed margin to costs. Even though this method covers the cost, it fails to account for demand elasticity, competitive actions, or supply fluctuations, making it unsuitable for dynamic market environments \cite{shafiee2010overview, li2015pricing}. Static pricing strategies have also been applied in energy and financial markets, where fixed prices are used for long-term contracts and auction-based transactions \cite{weron2014electricity}. While effective in stable markets, these models struggle in environments with high demand uncertainty or stochastic supply features. Recent research suggests that static pricing can sometimes achieve performance levels similar to dynamic pricing under specific conditions \cite{sun2024static}. 

\subsubsection{Dynamic Pricing}
Dynamic pricing models are being increasingly used across industries by adjusting prices in real-time based on demand fluctuations, inventory levels, and competitive behaviour \cite{hwang2006dynamic, el2023machine}. Methods such as Linear Regression, Random Forests, and Gradient Boosting are commonly used to forecast sales or estimate demand \cite{das2024optimizing, gupta2014machine}. Unsupervised learning methods, like clustering, support segmentation-based pricing strategies. However, these approaches usually fall short in adaptability to changes and may optimize for short-term gains without thinking about long-term customer behaviour. Real-world applications for this include airline ticket pricing, hotel bookings, and e-commerce platforms, where prices change dynamically. Despite their effectiveness, dynamic pricing strategies face challenges related to customer trust, market volatility, and complexity. As a result, there is growing interest in RL techniques that can learn pricing policies through interaction and adapt over time by explicitly modelling long-term objectives.

\subsection{Reinforcement Learning for Pricing Optimization}
Reinforcement learning has been heavily applied in decision-making tasks, including those of pricing optimization. Traditional RL methods, such as Q-learning, Deep Q-Networks (DQN), and Actor-Critic (A2C) algorithms, extend the Markov Decision Process (MDP) framework and have shown promising results in adaptive decision-making \cite{watkins1992q, roderick2017implementing, konda2003actor}. This section reviews two main approaches in RL for pricing: (1) Single-Agent Reinforcement Learning, where a single decision-maker optimizes its pricing strategy independently, and (2) Multi-Agent Reinforcement Learning (MARL), which extends RL to multiple interacting agents.  

\subsubsection{Single-Agent Reinforcement Learning}
Single-agent RL (SARL) has been extensively used in various decision-making and optimization tasks \cite{watkins1992q}. In SARL, an individual agent learns an optimal policy by interacting with an environment and receiving feedback in the form of rewards. The agent's goal is to maximize its expected reward over time. Traditional SARL methods rely on MDPs, assuming a stationary environment where the transition probabilities between states stay fixed \cite{hilton2023scaling}. The most fundamental RL algorithm, Q-learning, is a value-based method where an agent updates a Q-table to estimate the expected future reward for each state-action pair. However, this method struggles with high-dimensional state spaces, leading to the invention of DQNs. These use neural networks to approximate Q-values, enabling better generalization \cite{watkins1992q}. In addition to value-based approaches, policy gradient methods optimize the policy function by calculating gradients of the expected reward for the policy parameters. A2C methods combine value and policy-based learning, where the actor learns the policy and the critic evaluates the value function to improve learning stability \cite{georgila2014single}. These methods are particularly useful for continuous action spaces. Q-learning and DQN models struggle here because of discretization challenges. Recent research has explored the scalability of SARL models, showing that RL algorithms' performance improvements follow power-law scaling with model size \cite{hilton2023scaling}. This suggests that larger and more complex RL models can achieve better results. However, SARL faces significant challenges in dynamic and multi-agent environments by assuming a stationary and predictable environment. This is unrealistic in competitive markets and supply chain systems, where multiple decision-makers interact and all influence outcomes \cite{georgila2014single}.

\subsubsection{Multi-Agent Reinforcement Learning}
MARL extends regular RL by enabling multiple agents (such as manufacturers, suppliers, retailers, or robotic entities) to learn collaborative or competitive pricing strategies \cite{busoniu2008comprehensive}. MARL has been applied successfully to logistics in the supply chain, inventory management, vehicle routing, and demand forecasting. This is not well explored in dynamic pricing \cite{ren2022multi, kwon2001multi}. While single-agent RL assumes an isolated solitary decision-making entity to optimize its own strategy, MARL enables multiple agents to learn and make adaptive pricing decisions simultaneously with shared or competing objectives. This leads each agent to optimize its policy considering other evolving strategies, and hence a more dynamic and interactive environment for pricing. The interaction is most important in supply chains since the pricing, demand, inventory levels, and supply constraints are at different points of several interdependent entities \cite{blos2015modeling}. MARL's ability to capture strategic interactions makes it well-suited for dynamic pricing scenarios where traditional methods fail to account for these multi-agent dependencies.

\subsection{Challenges in MARL}
Despite its promise, MARL faces several challenges that have limited its application in complex practical settings. The decentralized interaction of agents leads to high computational overhead, problems with scalability, and instability of the training process. Other challenges in applications from supply chain management arise where the unavailability of quality training data and effective communication mechanisms among the agents are critical issues. A combination of algorithmic advancements and computational optimizations is needed to overcome these challenges. The following subsections highlight two major obstacles: Computational Complexity and Scalability and Data Limitations, together with potential solutions that mitigate these issues.

\subsubsection{Computational Complexity}
\label{computational_complexity}
MARL significantly increases computational demands due to the exponential growth of the joint state-action space as agents are added. Unlike SARL, where policy updates are independent, MARL agents must adapt to each others changing strategies, complicating optimization. This problem in dimensionality clogs exploration and leads to increasing training times \cite{zhou2023multi, wong2023deep}. A common solution is centralized training with decentralized execution (CTDE), where agents share information during training but act independently during execution. CTDE improves learning, but it provokes a high computational cost. Alternative strategies include mean-field approximations and graph-based learning. These reduce complexity by modelling only the local interactions \cite{wong2023deep}. To ease constraints of resources, parallel and distributed learning could help, though synchronizing policies in non-stationary settings remains a burden.

\subsubsection{Scalability and Data Limitations}
Scaling multi-agent systems worsens state-action growth, making convergence harder and instability better, particularly in dynamic supply chains where agents must learn competing agents' behaviour. Graph-based MARL and hierarchical learning are approaches to structuring interactions more efficiently \cite{wong2023deep}. Data scarcity is another challenge: ERP pricing data is often private or hidden, which makes research lean toward synthetic or simulated environments that may not capture real-world complexity. As a result, generalization across market conditions remains a key challenge \cite{zhou2023multi}. Also, inter-agent communication creates overhead, thus hindering training speed. Solutions include attention mechanisms and selective message filtering, which can improve communication efficiency without slowing down performance \cite{wong2023deep}.

\section{Methodology}
\label{sec:methodology}
This section describes the methodological framework developed to investigate how autonomous pricing agents can learn competitive strategies in a supply chain. The main task is to simulate a market environment in which pricing decisions from agents influence customer demand and the market. The core goal is to evaluate how different rule-based and MARL pricing strategies perform in terms of revenue, price stability, fairness, and market efficiency.

To do this, a real-world dataset was explored, preprocessed, and used to extract useful features in pricing and demand through feature selection and engineering. These features were used to train a predictive model that estimates customer demand with respect to price changes. This model functioned as the backbone for a custom simulation environment where agents set prices and receive feedback in a weekly cycle. The environment supported learning agents along with rule-based baselines. This was done by comparing strategic behaviours. The remainder of this section outlines the dataset selection and preparation, the construction of the demand model, the simulation setup, the agent architectures, and the evaluation protocol.

\subsection{Dataset Exploration and Justification}
To simulate realistic supply chain pricing and demand scenarios, this research made use of the publicly available Online Retail II dataset, released through the UCI Machine Learning Repository \cite{chen2012online}. The dataset contains over one million transactions recorded by a UK-based online retailer, specialized in giftware, serving both individual consumers and wholesalers, between December 2009 and December 2011. 

\subsubsection{Usage in studies}
Several studies have used this dataset to model customer segmentation and profitability dynamics. For instance, Chen et al. \cite{chen2012data} employed a Recency, Frequency, and Monetary (RFM) model through k-means clustering and decision tree induction to segment customers based on behaviour patterns. This confirms the suitability of the dataset for consumer-centric business intelligence implementations. In more recent work, Chen et al. \cite{chen2015predicting} continued to use this dataset to demonstrate the effectiveness of using RFM time series to model and predict customer profitability using a multilayer feed-forward neural network (MFNN). Also, the dataset's application extends to pattern mining. Singh et al. \cite{singh2018prefix} referenced real-world retail datasets, including the Online Retail II dataset, to validate algorithms to extract behavioural sequences that begin or end with specific events using prefix/suffix sequential pattern mining.

\subsubsection{Dataset characteristics}
The dataset's granularity and scale enabled detailed pricing and demand modelling across time, products, and customers. It covers 5,243 unique products and 5,876 customers. Each transaction included key variables such as product ID, timestamp, price, quantity, and customer ID, allowing for temporal, behavioural, and price-level analysis. An overview of the entire dataset structure, including column names, data types, and definitions, is provided in Appendix~\ref{sec:apx:dataset_description}.

\subsubsection{Exploratory Data Analysis}
EDA revealed several patterns relevant to pricing and demand modelling. B2B transactions were selected due to the presence of customer identifiers and notable behavioural differences; as non-registered customers paid on average 55.7\% more. Transactions were highly concentrated in the UK (92.1\%) and showed strong temporal effects, such as end-of-year seasonal peaks, midday spikes, and higher activity on Thursdays. Product pricing was highly variable, with some items showing volume discounts of over 50\%. Price volatility (as measured by coefficient of variation) averaged around 0.39, with a small subset exceeding 2.0 (as seen in Figure \ref{fig:price_volatility_distribution_appendix}) \cite{hendricks1936sampling}. Cancellations and returns were rare but were removed during cleaning. Complete descriptive statistics, volatility figures, and formulas are detailed in Appendix~\ref{sec:apx:eda}.

\subsection{Preprocessing}
\label{sec:preprocessing}
This section outlines the preprocessing pipeline applied to the dataset. This includes the removal of abnormal records, the extraction of customer and time-based features, and the construction of domain-specific variables. Additionally, product categories, week numbers, lagged and smoothed demand, price and trend indicators served as the foundational inputs for a demand prediction model as well as input states for agent models.

\subsubsection{Filtering and cleaning}
To ensure relevance and data quality, preprocessing started with the removal of invalid entries. Transactions with negative values for quantity or price were excluded, as well as records without a customer ID. These filters removed returns, errors, or overall incomplete purchases.

\subsubsection{Feature engineering}
To enhance predictive power and help align with the temporal structure of demand modelling, a variety of domain-relevant features were engineered from the dataset. 

\paragraph{Datetime and Country Features}
Multiple datetime indicators were extracted from each timestamp, including year, month, day, week, weekday, hour, and two binary flags: one for weekends (to capture behavioural differences between weekdays and weekends) and one for the holiday season, which is set to true during calendar weeks 47 to 52 to denote end-of-year peak activity. To support geographic segmentation, the categorical country column was also encoded into a numerical code label variable.

\paragraph {Semantic product clusters}
To capture semantic similarities between products, clustering was applied to the description field. Each unique product description was embedded using Sentence-BERT (MiniLM-L6-v2), a model designed to produce semantically meaningful sentence embeddings suitable for tasks such as clustering and semantic search \cite{reimers2019sentencebert}. The resulting vector representations were grouped using K-Means clustering into 20 product categories, generating a new categorical feature. This number was chosen to balance category differentiation with avoiding fine-grained or overlapping product categories. This allowed for the representation of similar products to be grouped on semantic meaning instead of sparse identifiers, aligning with evidence that clustering based on extracted feature representations can enhance modelling performance \cite{abdalgader2024experimental, bandara2020forecasting}.

\paragraph{Temporal demand signals}
Weekly aggregations were computed per product to align the dataset with the temporal structure of demand modelling. To capture short- and long term trends, several time-based features such as previous-week sales and over 2- and 4-week window rolling averages were derived \cite{bentaieb2012review}.

\paragraph{Transformations and interaction features}
To improve model behaviour and reduce skewness, demand was log-transformed. To also capture non-linear effects, price received both logarithmic and quadratic transformations. Seasonality was approximated using sine and cosine decompositions of calendar week and month values. Interaction terms (such as price-holiday, and price-category) allowed for conditional effects. Finally, features like momentum, trajectory, volatility, and price vs. category average were included to enhance sensitivity to recent temporal market behaviour. Definitions and equations for all indicators and features are in Table~\ref{tab:preprocessing_formulas}, Appendix~\ref{sec:apx:preprocessing_formulas}.

\subsection{Predicting Demand}
Forecasting weekly product demand is essential for simulating market behaviour and informing agent pricing. This section describes the supervised model used to simulate demand.

\subsubsection{LightGBM}
\label{sec:LightGBM}
To forecast weekly demand, a LightGBM (Light Gradient Boosting Machine) model \cite{ke2017lightgbm} was used. It was selected for its computational efficiency, scalability, and ability to handle high-dimensional tabular data with minimal preprocessing \cite{shwartz2022tabular}. Unlike deep learning models, LightGBM offers interpretability via feature importance scores, aligning with ERP requirements for explainability. The model was trained on a product-week aggregated dataset (8,777 observations, 21 features) with log-transformed demand as the target.

\begin{table}[h]
\centering
\caption{LightGBM Configuration and Performance Summary}
\label{tab:light_gbm}
\resizebox{\columnwidth}{!}{%
\begin{tabular}{ll}
\toprule
\textbf{Hyperparameter} & \textbf{Value} \\
\midrule
\texttt{n\_estimators}   & 2048 \\
\texttt{learning\_rate}  & 0.03 \\
\texttt{num\_leaves}     & 256 \\
Early stopping           & Patience = 100 on 10\% validation split \\
\midrule
\textbf{Cross-Validation $R^2$} & 0.591 $\pm$ 0.093 \\
\midrule
\textbf{Validation RMSE (log)} & 0.4957 \\
\textbf{Test RMSE (exp.)}   & 103.72 \\
\textbf{Test MAE (exp.)}    & 63.45 \\
\textbf{Test $R^2$ (exp.)}  & 0.74 \\
\bottomrule
\end{tabular}
}
\end{table}

As shown in Table~\ref{tab:light_gbm}, the model generalized well, achieving $R^2 = 0.74$ on the test set. The gap with cross-validation ($R^2 = 0.591 \pm 0.093$) suggests the test set contains more predictable patterns. This performance is sufficient to simulate realistic agent interactions.

\subsubsection{Feature Importance and Price Sensitivity}
\begin{figure}[H]
    \centering
    \includegraphics[width=0.48\textwidth]{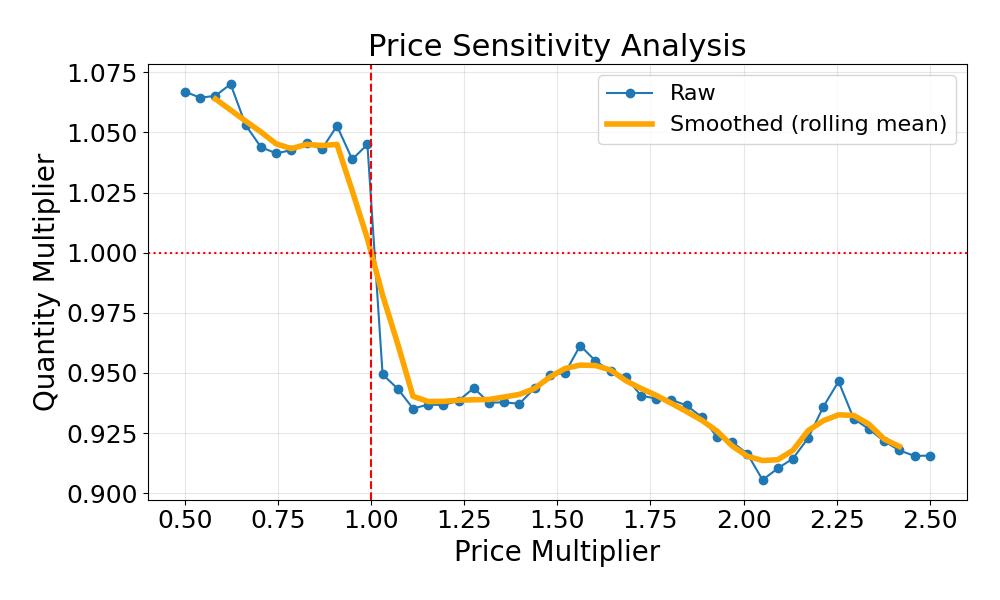}
    \caption{Smoothed price–demand curve}
    \label{fig:price_sensitivity}
\end{figure}

Temporal and pricing features, including lag, rolling means, trend, and volatility, ranked highest in feature importance (Appendix~\ref{sec:apx:preprocessing_formulas}), confirming their relevance. To assess price responsiveness, counterfactual analysis was performed by scaling price in the test set (0.5$\times$ to 2.5$\times$) while holding other inputs constant. The resulting demand curve (Figure~\ref{fig:price_sensitivity}) was smoothed using a 5-point centered rolling mean.

Price elasticity of demand ($\varepsilon$) was calculated as:
\begin{equation}
\varepsilon = \frac{\Delta Q / Q}{\Delta P / P}
\end{equation}

The resulting $\varepsilon = -0.072$ indicates inelastic demand, which is typical for giftware products \cite{wang2018modeling}. Although trained on log-transformed demand, elasticity was computed in the original scale using inverse-transformed demand predictions.

\subsection{Simulation Environment Design}
A custom simulation environment was developed to simulate market interactions and evaluate dynamic pricing strategies in a competitive setting. This environment provided a weekly time-stepped framework in which multiple pricing agents interact in a shared market and observe demand feedback influenced by their own and competitors' pricing decisions.

\subsubsection{Environment Overview}
The environment was initialized with a list of agents and a pre-trained LightGBM-based demand prediction model (see Section \ref{sec:LightGBM}). Each simulation step corresponds to one calendar week and updates the internal state variables such as the current date, year, week number, and holiday indicator. Agents submit product prices, and the environment uses the demand model to simulate weekly sales based on product-level features and competitive market conditions.

\subsubsection{State Sharing and Agent Feedback}
After predicting demand and calculating profits for each agent, the environment updates product-level demand histories and aggregates outcomes into a centralized history log. In addition, a structured market observation dictionary with week-number, holiday, and category clusters is compiled and passed to agents helping inform future pricing decisions.

\subsubsection{Agent Actions and Feedback Loop}
Each agent maintains a portfolio of products, each with its own cost structure and demand history. At each simulation step, agents observe the environment and select product pricing actions. For each step in the loop, it predicts demand, computes revenue, updates histories, and invokes an act on each agent to determine the next pricing action, allowing agents to refine their policies based on observed outcomes and evolving market conditions.

\subsection{MARL Agents}
Three distinct Multi-Agent Reinforcement Learning (MARL) frameworks were implemented and evaluated for optimal pricing strategies: Multi-Agent Deep Deterministic Policy Gradient (MADDPG), Multi-Agent Deep Q-Network (MADQN), and QMIX. Each offers different capabilities for the dynamic pricing domain, including continuous vs. discrete action spaces, degree of centralization, and agent coordination mechanisms.

\subsubsection{MADDPG}
Multi-Agent Deep Deterministic Policy Gradient (MADDPG) extends DDPG to multi-agent settings, enabling stable learning in non-stationary environments through centralized training and decentralized execution \cite{lowe2017multi}. Adapted from OpenAI's reference code for TensorFlow 2.x and dynamic pricing \cite{openai_maddpg}, each agent uses a local actor network to map observations to continuous pricing actions. A centralized critic evaluates joint actions using the global state. The full architecture of the MADDPG agent, including actor-critic structure, smoothing, and reward design, is illustrated in Appendix~\ref{sec:apx:maddpg_architecture}. This highlights the actor networks, action execution, centralized critics, and experience replay buffers, including custom contributions like recency-biased sampling and price stability penalties. The MADDPG implementation consists of:

\begin{enumerate}
    \item \textbf{Actor Network}: A 3-layer fully-connected policy network (ReLU activations, tanh output) mapping local observations to continuous pricing actions.
    
    \item \textbf{Critic Network}: A centralized architecture estimating Q-values based on joint observations and actions of all agents.
    
    \item \textbf{Experience Replay Buffer}: Stores state-action-reward transitions with a recency bias for prioritized sampling.
    
    \item \textbf{Exploration Noise}: Adds decaying Gaussian noise to the actor's output to encourage continuous exploration during training.
\end{enumerate}

While action decisions are decentralized, centralized critics leverage joint state-action information during training to improve learning stability across agents, encoding competitive market dynamics via price ratios, demand trends, seasonality, and market share metrics in agent state representations. During execution, each agent independently observes the market and adjusts prices by applying its actor network, but during training, the centralized critics leverage global information to compute more accurate gradients, facilitating stable policy learning across all agents.

\subsubsection{MADQN}

The Multi-Agent Deep Q-Network (MADQN) adapts DQN for multi-agent environments using discrete pricing actions ($-10\%$ to $+10\%$) \cite{mnih2015dqn, foerster2016learning}. Unlike MADDPG’s continuous control and centralized critics, MADQN agents learn independently, relying solely on local observations and rewards. This decentralization encourages self-interested behaviour, suitable for competitive pricing without coordination. MADQN tends to produce more aggressive strategies, leading to higher average prices and sharper differentiation between agents. Its architecture (Appendix~\ref{sec:apx:madqn_diagram}) includes a fully-connected Q-network (128, 64, 32 neurons), a target network for stability, $\varepsilon$-greedy exploration, and a recency-biased replay buffer to decorrelate updates. Custom elements such as price stability penalties and prioritized sampling distinguish this implementation from standard DQN \cite{li2021revisiting, liu2019dynamic}.

\subsubsection{QMIX}
QMIX represents a value-based MARL approach that overcomes the limitation of independent Q-learning (like MADQN) by enabling coordinated strategies. It combines per-agent utility functions (Q-values) into a single joint action-value function through a mixing network that enforces monotonicity constraints \cite{rashid2018qmix}.

Designed for decentralized execution with coordinated learning, each agent maintains its own Q-network for local action evaluation, while a centralized mixing network aggregates these per-agent utilities into a global action-value estimate. This mixing network, conditioned on a global state, ensures interpretability of local Q-values while enabling coordinated strategies. A coordinator pattern manages updates for efficient joint optimization during training.

QMIX's advantage lies in representing joint action-values which is not just a sum of individual functionalities, allowing it to learn policies where inter-product pricing coordination gives better returns than independent price optimization. As shown in Appendix~\ref{sec:apx:qmix_diagram}, its architecture includes agent-level Q-networks and a centralized mixing network. In practice, QMIX fosters a more balanced price-competition dynamic, discovering strategies that maintain competitive positioning and avoid destructive spirals, ensuring pricing decisions are locally informed yet globally coordinated.

\subsection{Baselines}
To contextualize MARL dynamic pricing, this study implements rule-based agents. Unlike learning agents, these do not adapt strategies, but follow fixed decision rules from common retail pricing practices. These rule-based agents serve as realistic, interpretable benchmarks, representing strategies typically embedded in ERP and pricing management systems.

\subsubsection{Rule-Based Pricing Agents}

Each Rule-Based agent follows a specific strategy that defines how prices are adjusted weekly. These strategies are deterministic but context-sensitive, reacting to competitor prices, historical data, demand signals, or seasonal effects. The following strategies were implemented:

\begin{enumerate}
\item \textbf{Static Markup}: A fixed cost-plus markup, independent of market conditions, serving as a simple baseline \cite{hinterhuber2004towards}.
\item \textbf{Competitor Matching}: Agent aligns price with the average competitor price in its category, with a small (1-5\%) undercut and a minimum margin \cite{chen2016empirical}.
\item \textbf{Historical Anchor}: Prices are set based on historical sales data, maintaining stability and resisting sudden market fluctuations \cite{nijs2007retail}.
\item \textbf{Demand Responsive}: Prices adjust dynamically to recent demand changes: up if demand increases, down if it declines, simulating agile data-driven responses to short-term sales \cite{wu2016automated}.
\item \textbf{Seasonal Pricing}: Prices are modulated by seasonal demand patterns, increasing during high-demand periods (such as holidays) and normalizing off-peak, exploiting predictable consumer activity spikes \cite{dong2019pricing}.
\end{enumerate}

\subsection{Experimental Setup}
The experimental framework was designed to evaluate the effectiveness of different pricing strategies in a competitive environment. A structured simulation was implemented, which allowed for controlled comparisons between MARL algorithms and baseline strategies.

\subsubsection{Simulation Environment Configuration}
The simulations were configured to model market dynamics over a span of two years, where each episode represented 104 weeks. At the start of each episode, four competing agents were initialized with an identical product portfolio. Each consisting of five products with category clusters (1, 2, 3, 5, and 10), where products were all initialized with identical starting prices. Then, the simulation was run for 30 episodes per experiment to allow sufficient time for MARL algorithms to converge.

\subsubsection{Algorithm Hyperparameters}
Each agent-type was carefully tuned and optimized for the retail pricing domain (see Table~\ref{tab:marl_hyperparameters} for an overview).

\begin{table}[h]
\centering
\caption{MARL Hyperparameters}
\label{tab:marl_hyperparameters}
\renewcommand{\arraystretch}{1.4}
\resizebox{\columnwidth}{!}{%
\begin{tabular}{llll}
\toprule
\textbf{Parameter} & \textbf{MADDPG} & \textbf{MADQN} & \textbf{QMIX} \\
\midrule
Learning rate & 0.0001 (actor) & 0.001 & 0.001 \\
              & 0.00001 (critic) &      &        \\
Discount factor & 0.95 & 0.95 & 0.95 \\
Exploration (decay) & Noise-based: 0.9995 & $\epsilon$-greedy: 0.995 & $\epsilon$-greedy: 0.995 \\
Target update & $\tau = 0.001$ & Every 5 steps & Every 5 steps \\
Batch size & 64 & 64 & 64 \\
\bottomrule
\end{tabular}
}
\end{table}

Exploration parameters were designed to decay exponentially with increasing episodes, facilitating thorough environment exploration in early training while progressively shifting toward exploitation. For MADDPG, exploration noise started at 0.2 and decayed multiplicatively by a factor of 0.9995 per episode, with a minimum threshold of 0.05. MADQN and QMIX employed an $\epsilon$-greedy exploration strategy, starting at 1.0 and decaying by 0.995 per episode, also with a minimum exploration rate of 0.05.

\subsubsection{Evaluation Metrics}
\label{sec:evaluation_metrics}

To assess the effectiveness of the implemented MARL algorithms and benchmark strategies, multiple evaluation metrics were employed, capturing both economic performance and market dynamics. In particular, this study emphasizes emergent fairness,  and market stability, reflecting the research objective to explore not only revenue optimization but also the broader implications of agent behaviour on market outcomes. 

The metrics include revenue per agent, price stability, Nash Equilibrium Proximity \cite{nash1950equilibrium}, optimality gap, welfare fairness \cite{ceriani2012origins, atkinson1999contributions}, Jain's fairness Index \cite{jain1984quantitative}, market share evolution, and price volatility. Together, these metrics provide a comprehensive view of agent adaptation, competitive behaviour, and overall market outcomes. Formal definitions and computational formulas for each metric are detailed in Appendix~\ref{sec:apx:evaluation_metrics} (Table~\ref{tab:evaluation_metrics_formulas}).

\subsubsection{Computational Resources}
\label{sec:computational-resources}
All experiments were executed on the Snellius National Supercomputer, operated by SURF in the Netherlands \cite{surf_snellius}. The experiments were configured to run on CPU-only mode. Each simulation utilized 8 CPU cores and was allocated 16 GB of RAM. The models were implemented using TensorFlow 2.x, optimized for CPU operations. Each full simulation was allocated up to 24 hours of computation time. The most compute-intensive components were the neural network training operations for MADDPG and QMIX.

\subsection{Reproducibility}
Experiments ran under fixed computational conditions (as detailed in Section~\ref{sec:computational-resources}), controlled by set scripts to minimize variability between runs. All experiment configurations (agent types, hyperparameters, simulation setups) were defined in dictionaries, serialized as JSON, and executed in isolated, and timestamped directories to prevent cross-contamination. Models, key simulation outputs, and performance metrics were systematically stored in standardized CSV files, logs, and visualizations. Methodological consistency was ensured through fixed time horizons (104 weeks per episode, 30 episodes) and unified evaluation windows. Finally, all simulations used a single, pre-trained demand model, and agents were initialized with identical product portfolios, pricing, and cost structures. The simulation environment maintained consistency with fixed weekly progressions, uniform agent action protocols, and centralized demand modelling.

\section{Results}
\label{sec:results}


Pricing adaptability refers to the ability for an agent to adjust prices to market conditions over time, which are expressed as changes in pricing magnitude and frequency. While high adaptability can enable quick responses to demand or competition fluctuations, if it is unregulated, it could cause severe instability. We evaluate it using two custom metrics (defined in Appendix~\ref{sec:apx:evaluation_metrics}): adjustment magnitude (average absolute percentage price change between time steps) and adjustment frequency (proportion of changes exceeding a 1\% threshold). 

\begin{table}[htbp]
\centering
\caption{Mean adaptability and stability metrics by agent type. Bold values indicate best-performing agents.}
\label{tab:adaptability_summary}
\begin{tabularx}{\linewidth}{@{}X r r r r@{}}
\toprule
\textbf{Agent Type} & \textbf{Adj. Mag.} & \textbf{Adj. Freq.} & \textbf{Stabil.} & \textbf{Volatil.} \\
\cmidrule[0.4pt](r{0.125em}){1-1}
\cmidrule[0.4pt](lr{0.125em}){2-2}
\cmidrule[0.4pt](lr{0.125em}){3-3}
\cmidrule[0.4pt](lr{0.125em}){4-4}
\cmidrule[0.4pt](l{0.25em}){5-5}
MADDPG & 0.0152 & 0.423 & 0.985 & 0.052 \\
\myrowcolour
MADQN & \textbf{0.0358} & \textbf{0.763} & 0.985 & 0.085 \\
MADDPG+MADQN & 0.0235 & 0.573 & 0.998 & 0.069 \\
\myrowcolour
MADDPG+QMIX & 0.0201 & 0.487 & 0.982 & 0.068 \\
MADDPG+Rule & 0.0127 & 0.340 & \textbf{1.000} & 0.050 \\
\myrowcolour
MADQN+Rule & 0.0339 & 0.771 & 0.996 & 0.083 \\
QMIX & 0.0243 & 0.553 & 0.972 & 0.075 \\
\myrowcolour
Rule-Based & 0.0114 & 0.312 & 0.944 & \textbf{0.024} \\
\bottomrule
\end{tabularx}
\end{table}

As shown in Table~\ref{tab:adaptability_summary}, MADQN agents demonstrate the most aggressive pricing, exhibiting the highest adjustment magnitude (\textit{0.0358}) and frequency (\textit{0.763}). This indicates a reactive, and exploratory strategy. QMIX and mixed MADQN configurations also show increased responsiveness to these metrics. Alternatively, MADDPG agents are more conservative (\textit{mean magnitude 0.0152}), with smoother and less volatile policy updates. As expected, rule-based agents have the lowest adjustment metrics due to their static design. Crucially, this means that high adaptability does not always lead to instability. Mixed strategies (such as MADDPG + Rule, MADDPG + MADQN) maintain high price stability (\textit{>0.98}) despite frequent adjustments. This suggests that hybrid agents can balance dynamic pricing with coordinated stability. However, MADQN's high volatility (\textit{0.085}) together with high adaptability implies that aggressive pricing can cause instability. Rule-based agents are the most stable (\textit{0.944-1.000}) and least volatile (\textit{0.024}), but mostly lack adaptability.

These findings highlight a trade-off: adaptability improves market responsiveness but it can generate instability unless balanced by conservative agents. Hybrid agent configurations are promising for achieving dynamic pricing while maintaining stable.

This subsection compares the performance of MARL agents with rule-based pricing agents across all configurations. The evaluation focuses on mean revenue generation, and market share.

\begin{figure}[htbp]
\centering
\includegraphics[width=0.9\linewidth]{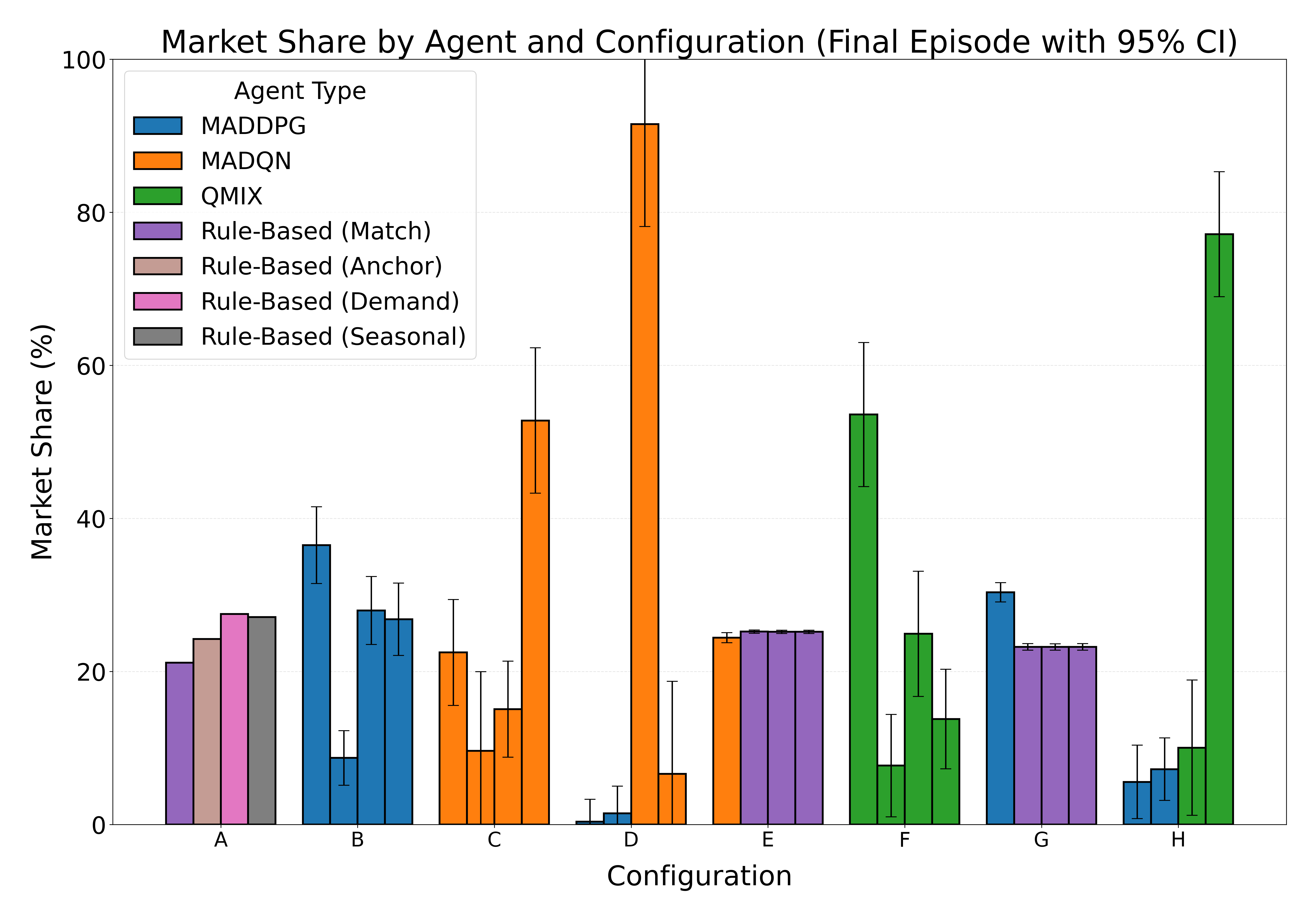}
\caption{Final Episode Market Share per Agent across Configurations (with 95\% CI).}
\label{fig:market_share_final}
\end{figure}

Figure~\ref{fig:market_share_final} illustrates the market share captured at the final episode for each agent across experimental configurations, including 95\% confidence intervals over each independent simulation run. This offers insight into long-term competitive dominance and strategic effectiveness.

To compare configurations statistically and isolate individual agent performance, Table~\ref{tab:mean_agent_returns} reports the mean return per agent, averaged across all episodes and eight independent simulation runs. This allows for a normalized comparison and highlights significant performance improvements over rule-based agents.

\begin{table}[htbp]
\centering
\caption{Mean agent return ($\bar{R}$) per configuration compared to Rule-Based baseline.}
\label{tab:mean_agent_returns}

\begin{tabularx}{\linewidth}{@{} X r r r @{}}
\toprule
\textbf{Configuration} & \textbf{$\bar{\textbf{R}}$ (\pounds)} & \textbf{$\Delta$} & \textbf{$\Delta$ (\%)} \\
\cmidrule[0.4pt](r{0.125em}){1-1}
\cmidrule[0.4pt](lr{0.125em}){2-2}
\cmidrule[0.4pt](lr{0.125em}){3-3}
\cmidrule[0.4pt](l{0.25em}){4-4}

A - All Rule-Based & \pounds22,817 & -- & -- \\
\myrowcolour
B - All MADDPG & \pounds89,861 & \pounds67,044 & +293.8\% \\
C - All MADQN & \pounds\textbf{997,669} & \pounds\textbf{974,852} & \textbf{+4272.5\%} \\
\myrowcolour
D - MADDPG + MADQN & \pounds709,224 & \pounds686,407 & +3008.3\% \\
E - MADQN + Rule-Based & \pounds945,045 & \pounds922,229 & +4041.9\% \\
\myrowcolour
F - All QMIX & \pounds393,121 & \pounds370,305 & +1622.9\% \\
G - One MADDPG & \pounds75,734 & \pounds52,917 & +231.9\% \\
\myrowcolour
H - MADDPG + QMIX & \pounds213,682 & \pounds190,865 & +836.5\% \\
\bottomrule
\end{tabularx}
\end{table}

To assess performance differences in per-agent reutrns, the Wilcoxon signed-rank test was applied. MARL-only configurations (B, C, F) were tested against the rule-based baseline (A). Even though consistent improvements were observed, statistical significance was not achieved $(p = 0.125)$. This was due to the small sample size (four agents). Future experiments with more agents could improve statistical power. Configurations with mixed types (D, E, G, H) were excluded from direct statistical testing.

\paragraph{Rule-based agents} achieve equitable market shares and highly stable pricing but generate substantially lower total revenue due to static pricing and limited responsiveness. Interestingly, in mixed configuration G (one MADDPG agent vs. three rule-based), total revenue modestly rises to \pounds628,016, suggesting rule-based agents can constrain isolated agents and that a single MARL agent is not enough to disrupt a stable rule-based market.

\paragraph{MADQN} agents exhibit the most pronounced revenue dominance, generating just under \pounds1 million in mean revenue through aggressive pricing. However, this comes with high price volatility and substantial market share fluctuations. This indicates significant short-term revenue generation but unstable competitive outcomes. This is probably due to a lack of coordinated pricing behaviour.

\paragraph{QMIX} agents consistently perform well in cooperative settings. They effectively coordinate pricing for strong revenue and sustained competitive positioning. Their performance deteriorates in heterogeneous configurations with learning assumptions that are conflicting. This highlights their strength in homogeneous or collaborative markets.

\paragraph{MADDPG} agents adopt a more cautious strategy. They achieve modest revenue and market share by prioritizing pricing stability over aggressive profit maximization. When combined with other agents (Configurations D and H), MADDPG contributes to dynamic outcomes but rarely dominates, which suggests conservative and risk-averse learning dynamics.

Overall, MARL agents outperform rule-based strategies in revenue generation but introduce trade-offs in volatility and coordination complexity. MADQN effectively brings value but can destabilize the market. QMIX offers a strong middle ground for cooperative agents, while MADDPG provides (suboptimal) stable pricing. Even though less adaptive, rule-based agents remain competitive in mixed environments. This proves that they are valuable in systems that prioritize transparency and stability.

To assess how market dynamics influence agent effectiveness, we analyse three key categories of outcomes. First, we examine emergent market dynamics, including fairness, and market volatility, which reflect structural conditions in the simulated environment. Second, we evaluate coordination behaviour through metrics such as Nash equilibrium proximity, and price convergence. Finally, we assess agent performance in economic terms via efficiency indicators, such as revenue optimality gap, and welfare fairness. These categories together allow us to analyse how different configurations influence total performance in terms of competitiveness, coordination, and responsiveness.

Table~\ref{tab:merged_metrics_summary} presents key indicators of market-level dynamics, coordination and performance. Fully rule-based environments (4x Rule) exhibit near-perfect fairness (\textit{0.9896} $\pm$ \textit{0.0000}), zero market share volatility, and low price instability. This forms a clear baseline for predictable dynamics. In contrast, configurations involving MADQN agents, exhibit significant volatility and lower fairness scores (such as \textit{0.5844} $\pm$ \textit{0.1625} for 4x MADQN). These patterns highlight a tendency for highly adaptive agents to disrupt equitable competition, introducing pricing instability. Hybrid configurations can either moderate or increase instability. For example, 1x MADQN + 3x Rule maintains high fairness (\textit{0.9991}) and low volatility (\textit{0.7pp}), indicating that the presence of a rule-based agent can constrain destabilizing behaviour. Alternatively, MADDPG + MADQN or MADDPG + QMIX configurations show lowered fairness and increased volatility. This suggests that hybridization does not necessarily stabilize dynamics without balance.

\begin{table*}[htbp]
\centering
\caption{Summary of emergent dynamics: Jain's Fairness (JF) and Market Volatility (MV), coordination: Nash Equilibrium Proximity (NEP) and Price Convergence (PC), and performance: Revenue Optimality Gap (ROG) and Welfare Fairness (WF) metrics across agent configurations.}
\label{tab:merged_metrics_summary}

\begin{tabularx}{\textwidth}{@{}
  >{\raggedright\arraybackslash}p{4.1cm}
  >{\centering\arraybackslash}p{2.93cm}
  >{\centering\arraybackslash}p{1.73cm}
  >{\centering\arraybackslash}p{1.73cm}
  >{\centering\arraybackslash}p{1.73cm}
  >{\centering\arraybackslash}p{1.73cm}
  >{\centering\arraybackslash}p{1.73cm}
@{}}
\toprule
\textbf{Agent Configuration} & \textbf{JF} & \textbf{MV (pp)} & \textbf{NEP} & \textbf{PC} & \textbf{ROG} & \textbf{WF} \\
\midrule
4x Rule (diverse strategies) & $0.9896 \pm 0.0000$ & \textbf{0.0} $\pm$ \textbf{0.0} & $0.8502$ & \textbf{0.9761} & $0.0000$ & $0.9451$ \\
\myrowcolour
4x MADDPG & $0.8819 \pm 0.1032$ & $9.5 \pm 1.8$ & $0.8639$ & $0.8745$ & \textbf{0.3957} & $0.9108$ \\
4x MADQN & $0.5844 \pm 0.1625$ & $22.4 \pm 5.2$ & $0.5788$ & $0.0131$ & $0.7060$ & $0.4383$ \\
\myrowcolour
2x MADDPG, 2x MADQN & $0.5290 \pm 0.1755$  & $16.9 \pm 12.2$ & $0.8782$ & $0.3698$ & $0.7480$ & $0.5894$ \\
1x MADQN, 3x Rule & $0.9991 \pm 0.0022$ & $0.7 \pm 0.4$ & $0.7396$ & $0.9237$ & $0.9596$ & \textbf{0.9908} \\
\myrowcolour
4x QMIX & $0.6953 \pm 0.1570$ & $17.5 \pm 1.8$ & $0.8071$ & $0.0402$ & $0.7272$ & $0.5146$ \\
1x MADDPG, 3x Rule & \textbf{0.9899} $\pm$ \textbf{0.0086} & $1.1 \pm 0.7$ & \textbf{0.9999} & $0.9324$ & $0.5355$ & $0.9921$ \\
\myrowcolour
2x MADDPG, 2x QMIX & $0.6417 \pm 0.2113$ & $16.4 \pm 8.3$ & $0.8406$ & $0.1329$ & $0.5432$ & $0.5279$ \\
\bottomrule
\end{tabularx}
\end{table*}

As shown in Table~\ref{tab:merged_metrics_summary}, strategic coordination varies considerably across configurations. Nash equilibrium proximity is highest in 1x MADDPG + 3x Rule (\textit{0.9999}) and 2x MADDPG + 2x MADQN (\textit{0.8782}). This indicates that hybrid configurations may encourage behaviour that seeks an equilibrium. This is potentially due to tension between adaptive and fixed strategies. Interestingly, 4x MADQN shows the lowest Nash equilibrium proximity (\textit{0.5788}). This strengthens earlier findings that these agents focus on aggressive and divergent strategies that usually disrupt stability. Price convergence follows similar trends: 4x Rule and 1x MADDPG + 3x Rule reach the highest convergence levels (\textit{0.9761} and \textit{0.9324}), while 4x MADQN yields almost no convergence at all.

Table~\ref{tab:merged_metrics_summary} examines whether dynamic strategies deliver economically meaningful improvements. Unsurprisingly, the lowest revenue optimality gap is observed in the rule-based configuration (\textit{0.0000}). Note that this does not come with high absolute performance. In contrast, 4x MADQN reaches a high gap (\textit{0.7060}), which suggests significant room between the actual and optimal pricing performance, even as it achieves the highest mean revenue overall. This confirms that aggressive pricing may exploit short-term gains without long-term efficiency. Finally, welfare fairness scores comply with earlier fairness metrics, with rule-based and hybrid configurations (such as 1x MADDPG + 3x Rule) maintaining high fairness. Conversely, MADQN and MADDPG + MADQN configurations rank lowest here, which suggests that price aggressiveness compromises not only fairness but also equitable welfare distribution.

These results highlight a fundamental feature in multi-agent learning environments. While adaptability can increase revenue, it often lowers market stability, fairness, and overall coordination. Rule-based and MADDPG agents deliver stable but less efficient outcomes. Conversely, MADQN excels in profit maximization at the cost of volatility and inequity. Hybrid configurations can strike a balance, but only when carefully configured and tuned. Agent interactions therefore shape not just outcomes but the very nature of the market itself. This makes configuration design a critical consideration in deploying such learning based pricing systems.

\section{Discussion}
\label{sec:discussion}

This section reflects on the key findings, examines how they relate to existing work, and considers the implications of the results. It also addresses limitations of the current study and touches on relevant ethical concerns.

\subsection{Comparison with State-of-the-Art}
Earlier studies on RL for pricing tend to focus on single-agent environments \cite{georgila2014single, neto2005single}. These approaches fall short when it comes to capturing the interactions between different supply chain actors. This study expands on this by testing several MARL methods: MADDPG, MADQN, and QMIX inside a market simulation based on actual retail data. It further advances recent MARL pricing research \cite{wong2023deep, zhou2023multi} through two contributions. First, it introduces hybrid agent populations, combining learning agents with rule-based strategies to reflect realistic market heterogeneity. Following, it evaluates performance using structural metrics such as fairness, coordination, and market stability, which remain under-explored in literature but are critical for practical deployment.

Among the tested models, MADQN stood out in terms of revenue generation. Its performance exceeded that of both static and single-agent baselines by a large margin. However, this level of gain came with costs which reflect similar patterns found in adversarial RL research \cite{lowe2017multi}. High revenues were often paired with instability and uneven market outcomes. In contrast, QMIX was better at promoting coordination among agents. Its behaviour aligns with findings in the cooperative MARL literature \cite{liang2025review}. MADDPG took a more stable path, often finding a trade-off between stability and adaptability. This study is set apart from purely simulated work due to the integration of a real-world demand model, which was built using LightGBM and enriched with price elasticity estimates.

\subsection{Interpretation of Findings}
The experiments highlight a trade-off between adaptability, fairness, and stability. MADQN agents adjusted prices most frequently and captured the highest revenue, but introduced volatility and inequity. Rule-based agents, by contrast, provided stability and fairness but lacked responsiveness, leading to lower revenue. Mixed agent configurations performed best overall. For instance, MADDPG paired with QMIX or rule-based agents achieved moderate adaptability while maintaining fairness and coordination. This suggests that combining exploratory and conservative agents balances market responsiveness and stability. Finally, MARL agents did not always dominate in mixed-agent settings. When surrounded by rule-based agents, even highly adaptive agents were limited in influence. This underscores the need to consider market composition and real-world constraints when deploying MARL in practice.

\subsection{Limitations}
While the results provide valuable insights into the behaviour of MARL agents in dynamic pricing scenarios, several limitations must be acknowledged to contextualize the findings and guide future research. The following sections outline key areas where these constraints were most notable.

\subsubsection{Realistic Price Elasticity}
The demand model produced a near-zero elasticity ($\varepsilon = -0.072$), which denotes inelastic behaviour typical of giftware sales, where demand is relatively insensitive to price changes. However, this is a limit in generalizability. In elastic markets like travel, digital services, or groceries, consumers respond more strongly to price shifts. Several MARL agents, including MADQN and QMIX, exploited this inelasticity by raising prices across episodes, knowing demand would remain fairly stable. While optimal in this context, such behaviour would likely be restricted in markets where elasticity burdens pricing power. This highlights the need for future simulations to use more elastic datasets or model heterogeneous consumer responses to avoid unrealistic, and loophole-driven strategies, ensuring that insights better reflect real-world dynamics.

\subsubsection{Product Portfolio}
Each agent handled a small, fixed group of products chosen from different clusters. While this setup allowed for controlled experiments, it does not reflect the complexity of actual supply chain environments. Real systems often involve large and nested product structures, cross-selling, and bundled pricing. Future work could explore agents that work over multiple levels of a catalogue and make combined decisions across related items.

\subsubsection{Bias in Data}
The dataset used for modelling comes from a single UK-based retailer, collected between 2009 and 2011. Even though it has a large transaction volume and structured format, making it suitable for simulation, several limitations affect generalizability. Customer behaviours, such as response to discounting, average order size, and purchasing frequency, could reflect UK consumer behaviour in the early 2010s. This feature heavily is shaped by local economic conditions and retail habits of that period. These patterns may be significantly different in other regions (such as North America or Asia) or in today’s e-commerce environments, where consumer expectations, promotional sensitivity, and platform dynamics have significantly changed. Similarly, seasonality patterns such as holiday spikes in demand may not align with global retail volatility and changes. As a result, the dataset supports controlled experimentation, but future studies should consider more recent and regionally diverse data sources, which would enhance realism and generalizability.

\subsubsection{Reproducibility and Scalability}
Reproducibility was a core principle in this study. This was ensured through deterministic data splits and consistent training procedures, but some variability in outcomes was observed due to inherent stochasticity of RL. Certain agents, such as MADQN and QMIX, occasionally exhibited different convergence behaviours across independent runs even with identical initial settings. Although this makes exact replication of individual trajectories challenging, average performance across runs was stable. These same agent types were particularly resource intensive, with training times of up to 12 hours per configuration on optimized 8-core CPU infrastructure (Snellius), reflecting the computational complexity discussed in Section~\ref{computational_complexity}. This limits scalability in larger markets or real-time ERP contexts. The current environment did not yet include other factors such as supply limits, delayed fulfilment, or inventory-sensitive pricing. Future work should explore how such constraints shape agent behaviour and influence pricing decisions.




\section{Conclusion}
\label{sec:conclusion}

This research explored how MARL can enhance dynamic pricing strategies in supply chains. This was done by addressing the gap between static pricing logic together with the need for adaptive, decentralized decision-making under market uncertainty. By benchmarking MADDPG, MADQN, and QMIX against rule-based pricing agents in a realistic market simulation, the study aimed to analyse and understand the trade-off between adaptability, revenue performance, fairness, and stability.

The main research question asked how MARL can improve dynamic pricing compared to traditional models while accounting for interactions between agents. The results showed that MARL, and MADQN in particular, significantly outperformed rule-based agents in terms of revenue, confirming the advantage of data-driven adaptation. However, this gain comes with a cost: higher volatility, reduced fairness, and coordination difficulties. This study showed that:

\begin{enumerate}
    \item MADQN was the most adaptive and aggressive, but destabilizing.
    \item Rule-based agents offered stability but were outperformed in revenue.
    \item Market dynamics shaped agent effectiveness significantly, with hybrid configurations balancing the trade-offs best.
\end{enumerate}

Concluding, while MARL has clear advantages, its effectiveness depends on the design of agent configurations and the underlying market conditions. Its benefits depend on how and where it is used, including algorithm setup and system limitations. The inelastic nature of demand in the dataset also limits generalizability to other markets. Nonetheless, the observed agent behaviours and trade-offs offer insights into how MARL systems behave under stable demand conditions. In particular, this study suggests that MARL is most beneficial when demand is predictable, competition is strategic, and price setting is flexible.

This study demonstrates that MARL can be practically integrated with ERP-relevant demand modelling. It also shows the importance of design considerations for dynamic pricing agents; particularly in low-elastic or regulated market segments. It also highlights that the environment and strategies of others strongly shape the performance of adaptive agents; an insight that is often missing from prior single-agent literature.

Future research should focus on two promising directions. First, it should incorporate products with elastic demand, which would help evaluate pricing responsiveness in more sensitive markets. Second, it should explore scalable MARL architectures, such as graph-based or hierarchical methods. These could improve coordination and training efficiency; especially in larger supply chains. These steps would help move MARL from experimental validation towards deployment readiness in business enterprise systems.

\bibliographystyle{ACM-Reference-Format}

\begin{thebibliography}{61}


\ifx \showCODEN    \undefined \def \showCODEN     #1{\unskip}     \fi
\ifx \showISBNx    \undefined \def \showISBNx     #1{\unskip}     \fi
\ifx \showISBNxiii \undefined \def \showISBNxiii  #1{\unskip}     \fi
\ifx \showISSN     \undefined \def \showISSN      #1{\unskip}     \fi
\ifx \showLCCN     \undefined \def \showLCCN      #1{\unskip}     \fi
\ifx \shownote     \undefined \def \shownote      #1{#1}          \fi
\ifx \showarticletitle \undefined \def \showarticletitle #1{#1}   \fi
\ifx \showURL      \undefined \def \showURL       {\relax}        \fi
\providecommand\bibfield[2]{#2}
\providecommand\bibinfo[2]{#2}
\providecommand\natexlab[1]{#1}
\providecommand\showeprint[2][]{arXiv:#2}

\bibitem[sur(2025)]%
        {surf_snellius}
 \bibinfo{year}{2025}\natexlab{}.
\newblock \bibinfo{title}{Snellius: de Nationale Supercomputer}.
\newblock
  \bibinfo{howpublished}{\url{https://www.surf.nl/snellius-de-nationale-supercomputer}}.
\newblock
\newblock
\shownote{Accessed: 2025-05-16}.


\bibitem[Abdalgader et~al\mbox{.}(2024)]%
        {abdalgader2024experimental}
\bibfield{author}{\bibinfo{person}{Khaled Abdalgader},
  \bibinfo{person}{Atheer~A Matroud}, {and} \bibinfo{person}{Khaled Hossin}.}
  \bibinfo{year}{2024}\natexlab{}.
\newblock \showarticletitle{Experimental study on short-text clustering using
  transformer-based semantic similarity measure}.
\newblock \bibinfo{journal}{\emph{PeerJ Computer Science}}
  \bibinfo{volume}{10} (\bibinfo{year}{2024}), \bibinfo{pages}{e2078}.
\newblock
\href{https://doi.org/10.7717/peerj-cs.2078}{doi:\nolinkurl{10.7717/peerj-cs.2078}}


\bibitem[Atkinson(1999)]%
        {atkinson1999contributions}
\bibfield{author}{\bibinfo{person}{Anthony~B Atkinson}.}
  \bibinfo{year}{1999}\natexlab{}.
\newblock \showarticletitle{The contributions of Amartya Sen to welfare
  economics}.
\newblock \bibinfo{journal}{\emph{The Scandinavian Journal of Economics}}
  \bibinfo{volume}{101}, \bibinfo{number}{2} (\bibinfo{year}{1999}),
  \bibinfo{pages}{173--190}.
\newblock
\href{https://doi.org/10.1111/1467-9442.00151}{doi:\nolinkurl{10.1111/1467-9442.00151}}


\bibitem[Bandara et~al\mbox{.}(2020)]%
        {bandara2020forecasting}
\bibfield{author}{\bibinfo{person}{Kasun Bandara}, \bibinfo{person}{Christoph
  Bergmeir}, {and} \bibinfo{person}{Slawek Smyl}.}
  \bibinfo{year}{2020}\natexlab{}.
\newblock \showarticletitle{Forecasting across time series databases using
  recurrent neural networks on groups of similar series: A clustering
  approach}.
\newblock \bibinfo{journal}{\emph{Expert Systems with Applications}}
  \bibinfo{volume}{140} (\bibinfo{year}{2020}), \bibinfo{pages}{112896}.
\newblock
\href{https://doi.org/10.1016/j.eswa.2019.112896}{doi:\nolinkurl{10.1016/j.eswa.2019.112896}}


\bibitem[Ben~Taieb et~al\mbox{.}(2012)]%
        {bentaieb2012review}
\bibfield{author}{\bibinfo{person}{Souhaib Ben~Taieb},
  \bibinfo{person}{Gianluca Bontempi}, \bibinfo{person}{Amir~F Atiya}, {and}
  \bibinfo{person}{Antti Sorjamaa}.} \bibinfo{year}{2012}\natexlab{}.
\newblock \showarticletitle{A review and comparison of strategies for
  multi-step ahead time series forecasting based on the NN5 forecasting
  competition}.
\newblock \bibinfo{journal}{\emph{Expert Systems with Applications}}
  \bibinfo{volume}{39}, \bibinfo{number}{8} (\bibinfo{year}{2012}),
  \bibinfo{pages}{7067--7083}.
\newblock
\href{https://doi.org/10.1016/j.eswa.2012.01.039}{doi:\nolinkurl{10.1016/j.eswa.2012.01.039}}


\bibitem[Blos and Miyagi(2015)]%
        {blos2015modeling}
\bibfield{author}{\bibinfo{person}{Maur{\'\i}cio~F Blos} {and}
  \bibinfo{person}{Paulo~E Miyagi}.} \bibinfo{year}{2015}\natexlab{}.
\newblock \showarticletitle{Modeling the supply chain disruptions: A study
  based on the supply chain interdependencies}.
\newblock \bibinfo{journal}{\emph{IFAC-PapersOnLine}} \bibinfo{volume}{48},
  \bibinfo{number}{3} (\bibinfo{year}{2015}), \bibinfo{pages}{2053--2058}.
\newblock
\href{https://doi.org/10.1016/j.ifacol.2015.06.391}{doi:\nolinkurl{10.1016/j.ifacol.2015.06.391}}


\bibitem[Busoniu et~al\mbox{.}(2008)]%
        {busoniu2008comprehensive}
\bibfield{author}{\bibinfo{person}{Lucian Busoniu}, \bibinfo{person}{Robert
  Babuska}, {and} \bibinfo{person}{Bart De~Schutter}.}
  \bibinfo{year}{2008}\natexlab{}.
\newblock \showarticletitle{A comprehensive survey of multiagent reinforcement
  learning}.
\newblock \bibinfo{journal}{\emph{IEEE Transactions on Systems, Man, and
  Cybernetics, Part C (Applications and Reviews)}} \bibinfo{volume}{38},
  \bibinfo{number}{2} (\bibinfo{year}{2008}), \bibinfo{pages}{156--172}.
\newblock
\href{https://doi.org/10.1109/TSMCC.2007.913919}{doi:\nolinkurl{10.1109/TSMCC.2007.913919}}


\bibitem[Cachon and Feldman(2010)]%
        {cachon2010dynamic}
\bibfield{author}{\bibinfo{person}{G{\'e}rard~P Cachon} {and}
  \bibinfo{person}{Pnina Feldman}.} \bibinfo{year}{2010}\natexlab{}.
\newblock \showarticletitle{Dynamic versus static pricing in the presence of
  strategic consumers}.
\newblock \bibinfo{journal}{\emph{The Wharton School, University of
  Pennsylvania}} (\bibinfo{year}{2010}).
\newblock


\bibitem[Ceriani and Verme(2012)]%
        {ceriani2012origins}
\bibfield{author}{\bibinfo{person}{Lidia Ceriani} {and} \bibinfo{person}{Paolo
  Verme}.} \bibinfo{year}{2012}\natexlab{}.
\newblock \showarticletitle{The origins of the Gini index: extracts from
  Variabilit{\`a} e Mutabilit{\`a} (1912) by Corrado Gini}.
\newblock \bibinfo{journal}{\emph{The Journal of Economic Inequality}}
  \bibinfo{volume}{10} (\bibinfo{year}{2012}), \bibinfo{pages}{421--443}.
\newblock
\href{https://doi.org/10.1007/s10888-011-9188-x}{doi:\nolinkurl{10.1007/s10888-011-9188-x}}


\bibitem[Chen(2012)]%
        {chen2012online}
\bibfield{author}{\bibinfo{person}{Daqing Chen}.}
  \bibinfo{year}{2012}\natexlab{}.
\newblock \bibinfo{title}{Online Retail II [Dataset]}.
\newblock
  \bibinfo{howpublished}{\url{https://archive.ics.uci.edu/dataset/502/online+retail+ii}}.
\newblock
\href{https://doi.org/10.24432/C5CG6D}{doi:\nolinkurl{10.24432/C5CG6D}}


\bibitem[Chen et~al\mbox{.}(2015)]%
        {chen2015predicting}
\bibfield{author}{\bibinfo{person}{Daqing Chen}, \bibinfo{person}{Kun Guo},
  {and} \bibinfo{person}{George Ubakanma}.} \bibinfo{year}{2015}\natexlab{}.
\newblock \showarticletitle{Predicting customer profitability over time based
  on RFM time series}.
\newblock \bibinfo{journal}{\emph{International Journal of Business Forecasting
  and Marketing Intelligence}} \bibinfo{volume}{2}, \bibinfo{number}{1}
  (\bibinfo{year}{2015}), \bibinfo{pages}{1--18}.
\newblock
\href{https://doi.org/10.1504/IJBFMI.2015.075325}{doi:\nolinkurl{10.1504/IJBFMI.2015.075325}}


\bibitem[Chen et~al\mbox{.}(2012)]%
        {chen2012data}
\bibfield{author}{\bibinfo{person}{Daqing Chen}, \bibinfo{person}{Sai~Laing
  Sain}, {and} \bibinfo{person}{Kun Guo}.} \bibinfo{year}{2012}\natexlab{}.
\newblock \showarticletitle{Data mining for the online retail industry: A case
  study of RFM model-based customer segmentation using data mining}.
\newblock \bibinfo{journal}{\emph{Journal of Database Marketing \& Customer
  Strategy Management}} \bibinfo{volume}{19}, \bibinfo{number}{3}
  (\bibinfo{year}{2012}), \bibinfo{pages}{197--208}.
\newblock
\href{https://doi.org/10.1057/dbm.2012.17}{doi:\nolinkurl{10.1057/dbm.2012.17}}


\bibitem[Chen et~al\mbox{.}(2016)]%
        {chen2016empirical}
\bibfield{author}{\bibinfo{person}{Le Chen}, \bibinfo{person}{Alan Mislove},
  {and} \bibinfo{person}{Christo Wilson}.} \bibinfo{year}{2016}\natexlab{}.
\newblock \showarticletitle{An Empirical Analysis of Algorithmic Pricing on
  Amazon Marketplace}. In \bibinfo{booktitle}{\emph{Proceedings of the 25th
  International Conference on World Wide Web}}. \bibinfo{publisher}{ACM},
  \bibinfo{pages}{1339--1349}.
\newblock
\href{https://doi.org/10.1145/2872427.2883089}{doi:\nolinkurl{10.1145/2872427.2883089}}


\bibitem[Chen et~al\mbox{.}(2021)]%
        {chen2021multiobjective}
\bibfield{author}{\bibinfo{person}{Yixian Chen}, \bibinfo{person}{Prakhar
  Mehrotra}, \bibinfo{person}{Nitin Kishore~Sai Samala},
  \bibinfo{person}{Kamilia Ahmadi}, \bibinfo{person}{Viresh Jivane},
  \bibinfo{person}{Linsey Pang}, \bibinfo{person}{Monika Shrivastav},
  \bibinfo{person}{Nate Lyman}, {and} \bibinfo{person}{Scott Pleiman}.}
  \bibinfo{year}{2021}\natexlab{}.
\newblock \showarticletitle{A Multiobjective Optimization for Clearance in
  Walmart Brick-and-Mortar Stores}.
\newblock \bibinfo{journal}{\emph{INFORMS Journal on Applied Analytics}}
  \bibinfo{volume}{51}, \bibinfo{number}{1} (\bibinfo{year}{2021}),
  \bibinfo{pages}{76--89}.
\newblock
\href{https://doi.org/10.1287/inte.2020.1065}{doi:\nolinkurl{10.1287/inte.2020.1065}}


\bibitem[Das et~al\mbox{.}(2024)]%
        {das2024optimizing}
\bibfield{author}{\bibinfo{person}{Pritom Das}, \bibinfo{person}{Tamanna
  Pervin}, \bibinfo{person}{Biswanath Bhattacharjee},
  \bibinfo{person}{Md~Razaul Karim}, \bibinfo{person}{Nasrin Sultana},
  \bibinfo{person}{Md~Sayham Khan}, \bibinfo{person}{Md~Afjal Hosien}, {and}
  \bibinfo{person}{FNU Kamruzzaman}.} \bibinfo{year}{2024}\natexlab{}.
\newblock \showarticletitle{Optimizing real-time dynamic pricing strategies in
  retail and e-commerce using machine learning models}.
\newblock \bibinfo{journal}{\emph{The American Journal of Engineering and
  Technology}} \bibinfo{volume}{6}, \bibinfo{number}{12}
  (\bibinfo{year}{2024}), \bibinfo{pages}{163--177}.
\newblock
\href{https://doi.org/10.37547/tajet/Volume06Issue12-15}{doi:\nolinkurl{10.37547/tajet/Volume06Issue12-15}}


\bibitem[Dong et~al\mbox{.}(2019)]%
        {dong2019pricing}
\bibfield{author}{\bibinfo{person}{Junfeng Dong}, \bibinfo{person}{Beilei Rao},
  \bibinfo{person}{Yu Liu}, \bibinfo{person}{Li Jiang},
  \bibinfo{person}{Wenxing Lu}, {and} \bibinfo{person}{Qiang Guo}.}
  \bibinfo{year}{2019}\natexlab{}.
\newblock \showarticletitle{Pricing strategies for different periods during
  subsequent selling season for seasonal products}.
\newblock \bibinfo{journal}{\emph{IEEE Access}}  \bibinfo{volume}{8}
  (\bibinfo{year}{2019}), \bibinfo{pages}{39479--39490}.
\newblock
\href{https://doi.org/10.1109/ACCESS.2019.2953284}{doi:\nolinkurl{10.1109/ACCESS.2019.2953284}}


\bibitem[El~Youbi et~al\mbox{.}(2023)]%
        {el2023machine}
\bibfield{author}{\bibinfo{person}{Raouya El~Youbi},
  \bibinfo{person}{Fay{\c{c}}al Messaoudi}, {and} \bibinfo{person}{Manal
  Loukili}.} \bibinfo{year}{2023}\natexlab{}.
\newblock \showarticletitle{Machine learning-driven dynamic pricing strategies
  in E-commerce}. In \bibinfo{booktitle}{\emph{2023 14th International
  Conference on Information and Communication Systems (ICICS)}}. IEEE,
  \bibinfo{pages}{1--5}.
\newblock
\href{https://doi.org/10.1109/ICICS60529.2023.10330541}{doi:\nolinkurl{10.1109/ICICS60529.2023.10330541}}


\bibitem[Emma(2024)]%
        {emma2024enterprise}
\bibfield{author}{\bibinfo{person}{Lawrence Emma}.}
  \bibinfo{year}{2024}\natexlab{}.
\newblock \showarticletitle{Enterprise Resource Planning (ERP) Systems for
  Streamlining Organizational Processes}.
\newblock \bibinfo{journal}{\emph{Unpublished Manuscript}}
  (\bibinfo{year}{2024}).
\newblock
\urldef\tempurl%
\url{https://www.researchgate.net/publication/386382658_Enterprise_Resource_Planning_ERP_Systems_for_Streamlining_Organizational_Processes}
\showURL{%
\tempurl}


\bibitem[Foerster et~al\mbox{.}(2016)]%
        {foerster2016learning}
\bibfield{author}{\bibinfo{person}{Jakob~N. Foerster},
  \bibinfo{person}{Yannis~M. Assael}, \bibinfo{person}{Nando de Freitas}, {and}
  \bibinfo{person}{Shimon Whiteson}.} \bibinfo{year}{2016}\natexlab{}.
\newblock \showarticletitle{Learning to Communicate with Deep Multi-Agent
  Reinforcement Learning}. In \bibinfo{booktitle}{\emph{Advances in Neural
  Information Processing Systems}}, Vol.~\bibinfo{volume}{29}.
  \bibinfo{publisher}{Curran Associates, Inc.}, \bibinfo{pages}{2137--2145}.
\newblock
\urldef\tempurl%
\url{https://proceedings.neurips.cc/paper/2016/file/c7635bfd99248a2cdef8249ef7bfbef4-Paper.pdf}
\showURL{%
\tempurl}
\newblock
\shownote{\url{https://arxiv.org/abs/1605.06676}}.


\bibitem[Georgila et~al\mbox{.}(2014)]%
        {georgila2014single}
\bibfield{author}{\bibinfo{person}{Kallirroi Georgila}, \bibinfo{person}{Claire
  Nelson}, {and} \bibinfo{person}{David Traum}.}
  \bibinfo{year}{2014}\natexlab{}.
\newblock \showarticletitle{Single-agent vs. multi-agent techniques for
  concurrent reinforcement learning of negotiation dialogue policies}. In
  \bibinfo{booktitle}{\emph{Proceedings of the 52nd Annual Meeting of the
  Association for Computational Linguistics (Volume 1: Long Papers)}}.
  \bibinfo{publisher}{Association for Computational Linguistics},
  \bibinfo{address}{Baltimore, Maryland}, \bibinfo{pages}{500--510}.
\newblock
\href{https://doi.org/10.3115/v1/P14-1047}{doi:\nolinkurl{10.3115/v1/P14-1047}}


\bibitem[Gupta and Pathak(2014)]%
        {gupta2014machine}
\bibfield{author}{\bibinfo{person}{Rajan Gupta} {and}
  \bibinfo{person}{Chaitanya Pathak}.} \bibinfo{year}{2014}\natexlab{}.
\newblock \showarticletitle{A machine learning framework for predicting
  purchase by online customers based on dynamic pricing}.
\newblock \bibinfo{journal}{\emph{Procedia Computer Science}}
  \bibinfo{volume}{36} (\bibinfo{year}{2014}), \bibinfo{pages}{599--605}.
\newblock
\href{https://doi.org/10.1016/j.procs.2014.09.060}{doi:\nolinkurl{10.1016/j.procs.2014.09.060}}


\bibitem[Hajji et~al\mbox{.}(2012)]%
        {hajji2012dynamic}
\bibfield{author}{\bibinfo{person}{Adn{\`e}ne Hajji}, \bibinfo{person}{Robert
  Pellerin}, \bibinfo{person}{Pierre-Majorique L{\'e}ger}, \bibinfo{person}{Ali
  Gharbi}, {and} \bibinfo{person}{Gilbert Babin}.}
  \bibinfo{year}{2012}\natexlab{}.
\newblock \showarticletitle{Dynamic pricing models for ERP systems under
  network externality}.
\newblock \bibinfo{journal}{\emph{International Journal of Production
  Economics}} \bibinfo{volume}{135}, \bibinfo{number}{2}
  (\bibinfo{year}{2012}), \bibinfo{pages}{708--715}.
\newblock
\href{https://doi.org/10.1016/j.ijpe.2011.10.004}{doi:\nolinkurl{10.1016/j.ijpe.2011.10.004}}


\bibitem[Han et~al\mbox{.}(2019)]%
        {han2019multi}
\bibfield{author}{\bibinfo{person}{Ye Han}, \bibinfo{person}{Xuefei Zhang},
  \bibinfo{person}{Jian Zhang}, \bibinfo{person}{Qimei Cui},
  \bibinfo{person}{Shuo Wang}, {and} \bibinfo{person}{Zhu Han}.}
  \bibinfo{year}{2019}\natexlab{}.
\newblock \showarticletitle{Multi-agent reinforcement learning enabling dynamic
  pricing policy for charging station operators}. In
  \bibinfo{booktitle}{\emph{2019 IEEE Global Communications Conference
  (GLOBECOM)}}. IEEE, \bibinfo{pages}{1--6}.
\newblock
\href{https://doi.org/10.1109/GLOBECOM38437.2019.9013999}{doi:\nolinkurl{10.1109/GLOBECOM38437.2019.9013999}}


\bibitem[Hendricks and Robey(1936)]%
        {hendricks1936sampling}
\bibfield{author}{\bibinfo{person}{Walter~A. Hendricks} {and}
  \bibinfo{person}{Kate~W. Robey}.} \bibinfo{year}{1936}\natexlab{}.
\newblock \showarticletitle{The Sampling Distribution of the Coefficient of
  Variation}.
\newblock \bibinfo{journal}{\emph{Annals of Mathematical Statistics}}
  \bibinfo{volume}{7}, \bibinfo{number}{3} (\bibinfo{year}{1936}),
  \bibinfo{pages}{129--132}.
\newblock
\href{https://doi.org/10.1214/aoms/1177732503}{doi:\nolinkurl{10.1214/aoms/1177732503}}


\bibitem[Hilton et~al\mbox{.}(2023)]%
        {hilton2023scaling}
\bibfield{author}{\bibinfo{person}{Jacob Hilton}, \bibinfo{person}{Jie Tang},
  {and} \bibinfo{person}{John Schulman}.} \bibinfo{year}{2023}\natexlab{}.
\newblock \showarticletitle{Scaling laws for single-agent reinforcement
  learning}.
\newblock \bibinfo{journal}{\emph{arXiv preprint arXiv:2301.13442}}
  (\bibinfo{year}{2023}).
\newblock
\href{https://doi.org/10.48550/arXiv.2301.13442}{doi:\nolinkurl{10.48550/arXiv.2301.13442}}


\bibitem[Hinterhuber(2004)]%
        {hinterhuber2004towards}
\bibfield{author}{\bibinfo{person}{Andreas Hinterhuber}.}
  \bibinfo{year}{2004}\natexlab{}.
\newblock \showarticletitle{Towards value-based pricing—An integrative
  framework for decision making}.
\newblock \bibinfo{journal}{\emph{Industrial Marketing Management}}
  \bibinfo{volume}{33}, \bibinfo{number}{8} (\bibinfo{year}{2004}),
  \bibinfo{pages}{765--778}.
\newblock
\href{https://doi.org/10.1016/j.indmarman.2003.10.006}{doi:\nolinkurl{10.1016/j.indmarman.2003.10.006}}


\bibitem[Hwang and Kim(2006)]%
        {hwang2006dynamic}
\bibfield{author}{\bibinfo{person}{Samuel~B Hwang} {and}
  \bibinfo{person}{Sungho Kim}.} \bibinfo{year}{2006}\natexlab{}.
\newblock \showarticletitle{Dynamic pricing algorithm for E-Commerce}. In
  \bibinfo{booktitle}{\emph{Advances in Systems, Computing Sciences and
  Software Engineering: Proceedings of SCSS05}}. \bibinfo{publisher}{Springer},
  \bibinfo{pages}{149--155}.
\newblock
\href{https://doi.org/10.1007/1-4020-5263-4_24}{doi:\nolinkurl{10.1007/1-4020-5263-4_24}}


\bibitem[Jain et~al\mbox{.}(1984)]%
        {jain1984quantitative}
\bibfield{author}{\bibinfo{person}{Rajendra~K Jain},
  \bibinfo{person}{Dah-Ming~W Chiu}, {and} \bibinfo{person}{William~R Hawe}.}
  \bibinfo{year}{1984}\natexlab{}.
\newblock \bibinfo{booktitle}{\emph{A Quantitative Measure of Fairness and
  Discrimination}}.
\newblock \bibinfo{type}{Technical Report} TR-301.
  \bibinfo{institution}{Digital Equipment Corporation},
  \bibinfo{address}{Hudson, MA}.
\newblock
\urldef\tempurl%
\url{https://www.cs.wustl.edu/~jain/papers/ftp/fairness.pdf}
\showURL{%
\tempurl}


\bibitem[Jain and Benyoucef(2008)]%
        {jain2008managing}
\bibfield{author}{\bibinfo{person}{Vipul Jain} {and} \bibinfo{person}{Lyes
  Benyoucef}.} \bibinfo{year}{2008}\natexlab{}.
\newblock \showarticletitle{Managing long supply chain networks: some emerging
  issues and challenges}.
\newblock \bibinfo{journal}{\emph{Journal of Manufacturing Technology
  Management}} \bibinfo{volume}{19}, \bibinfo{number}{4}
  (\bibinfo{year}{2008}), \bibinfo{pages}{469--496}.
\newblock
\href{https://doi.org/10.1108/17410380810869923}{doi:\nolinkurl{10.1108/17410380810869923}}


\bibitem[Ke et~al\mbox{.}(2017)]%
        {ke2017lightgbm}
\bibfield{author}{\bibinfo{person}{Guolin Ke}, \bibinfo{person}{Qi Meng},
  \bibinfo{person}{Thomas Finley}, \bibinfo{person}{Taifeng Wang},
  \bibinfo{person}{Wei Chen}, \bibinfo{person}{Weidong Ma},
  \bibinfo{person}{Qiwei Ye}, {and} \bibinfo{person}{Tie-Yan Liu}.}
  \bibinfo{year}{2017}\natexlab{}.
\newblock \showarticletitle{LightGBM: A Highly Efficient Gradient Boosting
  Decision Tree}. In \bibinfo{booktitle}{\emph{Advances in Neural Information
  Processing Systems}}, Vol.~\bibinfo{volume}{30}. \bibinfo{publisher}{Curran
  Associates, Inc.}, \bibinfo{pages}{3146--3154}.
\newblock
\urldef\tempurl%
\url{https://proceedings.neurips.cc/paper_files/paper/2017/file/6449f44a102fde848669bdd9eb6b76fa-Paper.pdf}
\showURL{%
\tempurl}


\bibitem[Kim et~al\mbox{.}(2015)]%
        {kim2015dynamic}
\bibfield{author}{\bibinfo{person}{Byung-Gook Kim}, \bibinfo{person}{Yu Zhang},
  \bibinfo{person}{Mihaela Van Der~Schaar}, {and} \bibinfo{person}{Jang-Won
  Lee}.} \bibinfo{year}{2015}\natexlab{}.
\newblock \showarticletitle{Dynamic pricing and energy consumption scheduling
  with reinforcement learning}.
\newblock \bibinfo{journal}{\emph{IEEE Transactions on Smart Grid}}
  \bibinfo{volume}{7}, \bibinfo{number}{5} (\bibinfo{year}{2015}),
  \bibinfo{pages}{2187--2198}.
\newblock
\href{https://doi.org/10.1109/TSG.2015.2495145}{doi:\nolinkurl{10.1109/TSG.2015.2495145}}


\bibitem[Konda and Tsitsiklis(2003)]%
        {konda2003actor}
\bibfield{author}{\bibinfo{person}{Vijay~R. Konda} {and}
  \bibinfo{person}{John~N. Tsitsiklis}.} \bibinfo{year}{2003}\natexlab{}.
\newblock \showarticletitle{On actor-critic algorithms}.
\newblock \bibinfo{journal}{\emph{SIAM Journal on Control and Optimization}}
  \bibinfo{volume}{42}, \bibinfo{number}{4} (\bibinfo{year}{2003}),
  \bibinfo{pages}{1143--1166}.
\newblock
\href{https://doi.org/10.1137/S0363012901385691}{doi:\nolinkurl{10.1137/S0363012901385691}}


\bibitem[Kwon and Lee(2001)]%
        {kwon2001multi}
\bibfield{author}{\bibinfo{person}{Oh~Byung Kwon} {and} \bibinfo{person}{J.J.
  Lee}.} \bibinfo{year}{2001}\natexlab{}.
\newblock \showarticletitle{A multi-agent intelligent system for efficient ERP
  maintenance}.
\newblock \bibinfo{journal}{\emph{Expert Systems with Applications}}
  \bibinfo{volume}{21}, \bibinfo{number}{4} (\bibinfo{year}{2001}),
  \bibinfo{pages}{191--202}.
\newblock
\href{https://doi.org/10.1016/S0957-4174(01)00039-2}{doi:\nolinkurl{10.1016/S0957-4174(01)00039-2}}


\bibitem[Li et~al\mbox{.}(2021)]%
        {li2021revisiting}
\bibfield{author}{\bibinfo{person}{Derek Li}, \bibinfo{person}{Andrew
  Jacobsen}, {and} \bibinfo{person}{Adam White}.}
  \bibinfo{year}{2021}\natexlab{}.
\newblock \showarticletitle{Revisiting Experience Replay in Non-Stationary
  Environments}. In \bibinfo{booktitle}{\emph{Proceedings of the Adaptive and
  Learning Agents Workshop (ALA)}}.
\newblock
\urldef\tempurl%
\url{https://ala2021.vub.ac.be/papers/ALA2021_paper_51.pdf}
\showURL{%
\tempurl}


\bibitem[Li et~al\mbox{.}(2015)]%
        {li2015pricing}
\bibfield{author}{\bibinfo{person}{Le Li}, \bibinfo{person}{Xiao Lin},
  \bibinfo{person}{Rudy~R Negenborn}, {and} \bibinfo{person}{Bart
  De~Schutter}.} \bibinfo{year}{2015}\natexlab{}.
\newblock \showarticletitle{Pricing intermodal freight transport services: A
  cost-plus-pricing strategy}. In \bibinfo{booktitle}{\emph{Computational
  Logistics: 6th International Conference, ICCL 2015, Delft, The Netherlands,
  September 23-25, 2015, Proceedings 6}}. Springer, \bibinfo{pages}{541--556}.
\newblock
\href{https://doi.org/10.1007/978-3-319-24264-4_37}{doi:\nolinkurl{10.1007/978-3-319-24264-4_37}}


\bibitem[Liang et~al\mbox{.}(2025)]%
        {liang2025review}
\bibfield{author}{\bibinfo{person}{Jiaxin Liang}, \bibinfo{person}{Haotian
  Miao}, \bibinfo{person}{Kai Li}, \bibinfo{person}{Jianheng Tan},
  \bibinfo{person}{Xi Wang}, \bibinfo{person}{Rui Luo}, {and}
  \bibinfo{person}{Yueqiu Jiang}.} \bibinfo{year}{2025}\natexlab{}.
\newblock \showarticletitle{A Review of Multi-Agent Reinforcement Learning
  Algorithms}.
\newblock \bibinfo{journal}{\emph{Electronics}} \bibinfo{volume}{14},
  \bibinfo{number}{4} (\bibinfo{year}{2025}), \bibinfo{pages}{820}.
\newblock
\href{https://doi.org/10.3390/electronics14040820}{doi:\nolinkurl{10.3390/electronics14040820}}


\bibitem[Liu et~al\mbox{.}(2019)]%
        {liu2019dynamic}
\bibfield{author}{\bibinfo{person}{Jiaxi Liu}, \bibinfo{person}{Yidong Zhang},
  \bibinfo{person}{Xiaoqing Wang}, \bibinfo{person}{Yuming Deng}, {and}
  \bibinfo{person}{Xingyu Wu}.} \bibinfo{year}{2019}\natexlab{}.
\newblock \showarticletitle{Dynamic Pricing on E-commerce Platform with Deep
  Reinforcement Learning: A Field Experiment}.
\newblock \bibinfo{journal}{\emph{arXiv preprint arXiv:1912.02572}}
  (\bibinfo{year}{2019}).
\newblock
\href{https://doi.org/10.48550/arXiv.1912.02572}{doi:\nolinkurl{10.48550/arXiv.1912.02572}}


\bibitem[Lowe et~al\mbox{.}(2017a)]%
        {lowe2017multi}
\bibfield{author}{\bibinfo{person}{Ryan Lowe}, \bibinfo{person}{Yi Wu},
  \bibinfo{person}{Aviv Tamar}, \bibinfo{person}{Jean Harb},
  \bibinfo{person}{Pieter Abbeel}, {and} \bibinfo{person}{Igor Mordatch}.}
  \bibinfo{year}{2017}\natexlab{a}.
\newblock \showarticletitle{Multi-Agent Actor-Critic for Mixed
  Cooperative-Competitive Environments}. In \bibinfo{booktitle}{\emph{Advances
  in Neural Information Processing Systems}}, Vol.~\bibinfo{volume}{30}.
  \bibinfo{publisher}{Curran Associates, Inc.}, \bibinfo{pages}{6379--6390}.
\newblock
\urldef\tempurl%
\url{https://proceedings.neurips.cc/paper/2017/file/68a9750337a418a86fe06c1991a1d64c-Paper.pdf}
\showURL{%
\tempurl}


\bibitem[Lowe et~al\mbox{.}(2017b)]%
        {openai_maddpg}
\bibfield{author}{\bibinfo{person}{Ryan Lowe}, \bibinfo{person}{Yi Wu},
  \bibinfo{person}{Aviv Tamar}, \bibinfo{person}{Jean Harb},
  \bibinfo{person}{Pieter Abbeel}, {and} \bibinfo{person}{Igor Mordatch}.}
  \bibinfo{year}{2017}\natexlab{b}.
\newblock \bibinfo{title}{Multi-Agent Deep Deterministic Policy Gradient
  (MADDPG) - GitHub repository}.
\newblock
\urldef\tempurl%
\url{https://github.com/openai/maddpg}
\showURL{%
\tempurl}
\newblock
\shownote{Archived codebase provided as-is, based on the NeurIPS 2017 paper}.


\bibitem[Menon(2024)]%
        {menon2024erp}
\bibfield{author}{\bibinfo{person}{Arun~K. Menon}.}
  \bibinfo{year}{2024}\natexlab{}.
\newblock \showarticletitle{Integrating Pricing Optimization Models with ERP
  Systems to Enhance Profitability in U.S. E-commerce Supply Chains}.
\newblock \bibinfo{journal}{\emph{Global Journal of Engineering and Technology
  Advances}} \bibinfo{volume}{20}, \bibinfo{number}{2} (\bibinfo{year}{2024}),
  \bibinfo{pages}{242--255}.
\newblock
\href{https://doi.org/10.30574/gjeta.2024.20.2.0151}{doi:\nolinkurl{10.30574/gjeta.2024.20.2.0151}}
\newblock
\shownote{Open access under CC BY 4.0}.


\bibitem[Mnih et~al\mbox{.}(2015)]%
        {mnih2015dqn}
\bibfield{author}{\bibinfo{person}{Volodymyr Mnih}, \bibinfo{person}{Koray
  Kavukcuoglu}, \bibinfo{person}{David Silver}, \bibinfo{person}{Andrei~A
  Rusu}, \bibinfo{person}{Joel Veness}, \bibinfo{person}{Marc~G Bellemare},
  \bibinfo{person}{Alex Graves}, \bibinfo{person}{Martin Riedmiller},
  \bibinfo{person}{Andreas~K Fidjeland}, \bibinfo{person}{Georg Ostrovski},
  \bibinfo{person}{Stig Petersen}, \bibinfo{person}{Charles Beattie},
  \bibinfo{person}{Amir Sadik}, \bibinfo{person}{Ioannis Antonoglou},
  \bibinfo{person}{Helen King}, \bibinfo{person}{Dharshan Kumaran},
  \bibinfo{person}{Daan Wierstra}, \bibinfo{person}{Shane Legg}, {and}
  \bibinfo{person}{Demis Hassabis}.} \bibinfo{year}{2015}\natexlab{}.
\newblock \showarticletitle{Human-level control through deep reinforcement
  learning}.
\newblock \bibinfo{journal}{\emph{Nature}} \bibinfo{volume}{518},
  \bibinfo{number}{7540} (\bibinfo{year}{2015}), \bibinfo{pages}{529--533}.
\newblock
\href{https://doi.org/10.1038/nature14236}{doi:\nolinkurl{10.1038/nature14236}}


\bibitem[Nash~Jr(1950)]%
        {nash1950equilibrium}
\bibfield{author}{\bibinfo{person}{John~F Nash~Jr}.}
  \bibinfo{year}{1950}\natexlab{}.
\newblock \showarticletitle{Equilibrium Points in N-Person Games}.
\newblock \bibinfo{journal}{\emph{Proceedings of the National Academy of
  Sciences}} \bibinfo{volume}{36}, \bibinfo{number}{1} (\bibinfo{year}{1950}),
  \bibinfo{pages}{48--49}.
\newblock
\href{https://doi.org/10.1073/pnas.36.1.48}{doi:\nolinkurl{10.1073/pnas.36.1.48}}


\bibitem[Neto(2005)]%
        {neto2005single}
\bibfield{author}{\bibinfo{person}{Gon{\c{c}}alo Neto}.}
  \bibinfo{year}{2005}\natexlab{}.
\newblock \bibinfo{title}{From Single-Agent to Multi-Agent Reinforcement
  Learning: Foundational Concepts and Methods}.
\newblock
\urldef\tempurl%
\url{https://users.cs.utah.edu/~tch/CS6380/resources/Neto-2005-RL-MAS-Tutorial.pdf}
\showURL{%
\tempurl}
\newblock
\shownote{Learning Theory Course 2005; 2}.


\bibitem[Nijs et~al\mbox{.}(2007)]%
        {nijs2007retail}
\bibfield{author}{\bibinfo{person}{Vincent~R Nijs}, \bibinfo{person}{Shuba
  Srinivasan}, {and} \bibinfo{person}{Koen Pauwels}.}
  \bibinfo{year}{2007}\natexlab{}.
\newblock \showarticletitle{Retail-price drivers and retailer profits}.
\newblock \bibinfo{journal}{\emph{Marketing Science}} \bibinfo{volume}{26},
  \bibinfo{number}{4} (\bibinfo{year}{2007}), \bibinfo{pages}{473--487}.
\newblock
\href{https://doi.org/10.1287/mksc.1060.0205}{doi:\nolinkurl{10.1287/mksc.1060.0205}}


\bibitem[Rashid et~al\mbox{.}(2018)]%
        {rashid2018qmix}
\bibfield{author}{\bibinfo{person}{Tabish Rashid}, \bibinfo{person}{Mikayel
  Samvelyan}, \bibinfo{person}{Christian Schroeder~de Witt},
  \bibinfo{person}{Gregory Farquhar}, \bibinfo{person}{Jakob Foerster}, {and}
  \bibinfo{person}{Shimon Whiteson}.} \bibinfo{year}{2018}\natexlab{}.
\newblock \showarticletitle{QMIX: Monotonic Value Function Factorisation for
  Deep Multi-Agent Reinforcement Learning}. In
  \bibinfo{booktitle}{\emph{Proceedings of the 35th International Conference on
  Machine Learning}}. PMLR, \bibinfo{pages}{4295--4304}.
\newblock
\href{https://doi.org/10.48550/arXiv.1803.11485}{doi:\nolinkurl{10.48550/arXiv.1803.11485}}


\bibitem[Reimers and Gurevych(2019)]%
        {reimers2019sentencebert}
\bibfield{author}{\bibinfo{person}{Nils Reimers} {and} \bibinfo{person}{Iryna
  Gurevych}.} \bibinfo{year}{2019}\natexlab{}.
\newblock \showarticletitle{Sentence-BERT: Sentence Embeddings using Siamese
  BERT-Networks}. In \bibinfo{booktitle}{\emph{Proceedings of the 2019
  Conference on Empirical Methods in Natural Language Processing and the 9th
  International Joint Conference on Natural Language Processing
  (EMNLP-IJCNLP)}}. Association for Computational Linguistics,
  \bibinfo{pages}{3982--3992}.
\newblock
\href{https://doi.org/10.18653/v1/D19-1410}{doi:\nolinkurl{10.18653/v1/D19-1410}}


\bibitem[Ren et~al\mbox{.}(2022)]%
        {ren2022multi}
\bibfield{author}{\bibinfo{person}{Lei Ren}, \bibinfo{person}{Xiaoyang Fan},
  \bibinfo{person}{Jin Cui}, \bibinfo{person}{Zhen Shen},
  \bibinfo{person}{Yisheng Lv}, {and} \bibinfo{person}{Gang Xiong}.}
  \bibinfo{year}{2022}\natexlab{}.
\newblock \showarticletitle{A multi-agent reinforcement learning method with
  route recorders for vehicle routing in supply chain management}.
\newblock \bibinfo{journal}{\emph{IEEE Transactions on Intelligent
  Transportation Systems}} \bibinfo{volume}{23}, \bibinfo{number}{9}
  (\bibinfo{year}{2022}), \bibinfo{pages}{16410--16420}.
\newblock
\href{https://doi.org/10.1109/TITS.2022.3150151}{doi:\nolinkurl{10.1109/TITS.2022.3150151}}


\bibitem[Roderick et~al\mbox{.}(2017)]%
        {roderick2017implementing}
\bibfield{author}{\bibinfo{person}{Melrose Roderick}, \bibinfo{person}{James
  MacGlashan}, {and} \bibinfo{person}{Stefanie Tellex}.}
  \bibinfo{year}{2017}\natexlab{}.
\newblock \showarticletitle{Implementing the deep q-network}.
\newblock \bibinfo{journal}{\emph{arXiv preprint arXiv:1711.07478}}
  (\bibinfo{year}{2017}).
\newblock
\href{https://doi.org/10.48550/arXiv.1711.07478}{doi:\nolinkurl{10.48550/arXiv.1711.07478}}


\bibitem[Shafiee and Topal(2010)]%
        {shafiee2010overview}
\bibfield{author}{\bibinfo{person}{Shahriar Shafiee} {and}
  \bibinfo{person}{Erkan Topal}.} \bibinfo{year}{2010}\natexlab{}.
\newblock \showarticletitle{An overview of global gold market and gold price
  forecasting}.
\newblock \bibinfo{journal}{\emph{Resources Policy}} \bibinfo{volume}{35},
  \bibinfo{number}{3} (\bibinfo{year}{2010}), \bibinfo{pages}{178--189}.
\newblock
\href{https://doi.org/10.1016/j.resourpol.2010.05.004}{doi:\nolinkurl{10.1016/j.resourpol.2010.05.004}}


\bibitem[Shwartz-Ziv and Armon(2022)]%
        {shwartz2022tabular}
\bibfield{author}{\bibinfo{person}{Ravid Shwartz-Ziv} {and}
  \bibinfo{person}{Amitai Armon}.} \bibinfo{year}{2022}\natexlab{}.
\newblock \showarticletitle{Tabular Data: Deep Learning is Not All You Need}.
\newblock \bibinfo{journal}{\emph{Information Fusion}}  \bibinfo{volume}{81}
  (\bibinfo{year}{2022}), \bibinfo{pages}{84--90}.
\newblock
\href{https://doi.org/10.1016/j.inffus.2021.11.011}{doi:\nolinkurl{10.1016/j.inffus.2021.11.011}}


\bibitem[Singh et~al\mbox{.}(2018)]%
        {singh2018prefix}
\bibfield{author}{\bibinfo{person}{Rina Singh}, \bibinfo{person}{Jeffrey~A.
  Graves}, \bibinfo{person}{Douglas~A. Talbert}, {and} \bibinfo{person}{William
  Eberle}.} \bibinfo{year}{2018}\natexlab{}.
\newblock \showarticletitle{Prefix and suffix sequential pattern mining}. In
  \bibinfo{booktitle}{\emph{Advances in Data Mining. Applications and
  Theoretical Aspects}}. \bibinfo{publisher}{Springer},
  \bibinfo{pages}{309--324}.
\newblock
\href{https://doi.org/10.1007/978-3-319-95786-9_24}{doi:\nolinkurl{10.1007/978-3-319-95786-9_24}}


\bibitem[Smith(2024)]%
        {smith2024dynamic}
\bibfield{author}{\bibinfo{person}{Hussein~Kamaldeen Smith}.}
  \bibinfo{year}{2024}\natexlab{}.
\newblock \showarticletitle{Dynamic Pricing and E-supply Chain Coordination:
  Effects on Inventory Optimization and Profit Margins}.
\newblock \bibinfo{journal}{\emph{Unpublished Manuscript}}
  (\bibinfo{year}{2024}).
\newblock
\urldef\tempurl%
\url{https://www.researchgate.net/publication/384805839_Dynamic_Pricing_and_E-supply_Chain_Coordination_Effects_on_Inventory_Optimization_and_Profit_Margins}
\showURL{%
\tempurl}


\bibitem[Sun et~al\mbox{.}(2024)]%
        {sun2024static}
\bibfield{author}{\bibinfo{person}{Bo Sun}, \bibinfo{person}{Hossein~Nekouyan
  Jazi}, \bibinfo{person}{Xiaoqi Tan}, {and} \bibinfo{person}{Raouf Boutaba}.}
  \bibinfo{year}{2024}\natexlab{}.
\newblock \showarticletitle{Static Pricing for Online Selection Problem and its
  Variants}.
\newblock \bibinfo{journal}{\emph{arXiv preprint arXiv:2410.07378}}
  (\bibinfo{year}{2024}).
\newblock
\href{https://doi.org/10.48550/arXiv.2410.07378}{doi:\nolinkurl{10.48550/arXiv.2410.07378}}


\bibitem[Tarn et~al\mbox{.}(2002)]%
        {tarn2002exploring}
\bibfield{author}{\bibinfo{person}{J~Michael Tarn}, \bibinfo{person}{David~C
  Yen}, {and} \bibinfo{person}{Marcus Beaumont}.}
  \bibinfo{year}{2002}\natexlab{}.
\newblock \showarticletitle{Exploring the rationales for ERP and SCM
  integration}.
\newblock \bibinfo{journal}{\emph{Industrial Management \& Data Systems}}
  \bibinfo{volume}{102}, \bibinfo{number}{1} (\bibinfo{year}{2002}),
  \bibinfo{pages}{26--34}.
\newblock
\href{https://doi.org/10.1108/02635570210414631}{doi:\nolinkurl{10.1108/02635570210414631}}


\bibitem[Wang et~al\mbox{.}(2023)]%
        {wang2023algorithms}
\bibfield{author}{\bibinfo{person}{Qiaochu Wang}, \bibinfo{person}{Yan Huang},
  \bibinfo{person}{Param~Vir Singh}, {and} \bibinfo{person}{Kannan
  Srinivasan}.} \bibinfo{year}{2023}\natexlab{}.
\newblock \showarticletitle{Algorithms, Artificial Intelligence and Simple Rule
  Based Pricing}.
\newblock \bibinfo{journal}{\emph{SSRN Electronic Journal}}
  (\bibinfo{year}{2023}).
\newblock
\href{https://doi.org/10.2139/ssrn.4144905}{doi:\nolinkurl{10.2139/ssrn.4144905}}


\bibitem[Wang and Van Der~Lans(2018)]%
        {wang2018modeling}
\bibfield{author}{\bibinfo{person}{Sherry~Shi Wang} {and} \bibinfo{person}{Ralf
  Van Der~Lans}.} \bibinfo{year}{2018}\natexlab{}.
\newblock \showarticletitle{Modeling gift choice: The effect of uncertainty on
  price sensitivity}.
\newblock \bibinfo{journal}{\emph{Journal of Marketing Research}}
  \bibinfo{volume}{55}, \bibinfo{number}{4} (\bibinfo{year}{2018}),
  \bibinfo{pages}{524--540}.
\newblock
\href{https://doi.org/10.1509/jmr.16.0453}{doi:\nolinkurl{10.1509/jmr.16.0453}}


\bibitem[Watkins and Dayan(1992)]%
        {watkins1992q}
\bibfield{author}{\bibinfo{person}{Christopher~JCH Watkins} {and}
  \bibinfo{person}{Peter Dayan}.} \bibinfo{year}{1992}\natexlab{}.
\newblock \showarticletitle{Q-learning}.
\newblock \bibinfo{journal}{\emph{Machine Learning}} \bibinfo{volume}{8},
  \bibinfo{number}{3} (\bibinfo{year}{1992}), \bibinfo{pages}{279--292}.
\newblock
\href{https://doi.org/10.1007/BF00992698}{doi:\nolinkurl{10.1007/BF00992698}}


\bibitem[Weron(2014)]%
        {weron2014electricity}
\bibfield{author}{\bibinfo{person}{Rafa{\l} Weron}.}
  \bibinfo{year}{2014}\natexlab{}.
\newblock \showarticletitle{Electricity price forecasting: A review of the
  state-of-the-art with a look into the future}.
\newblock \bibinfo{journal}{\emph{International Journal of Forecasting}}
  \bibinfo{volume}{30}, \bibinfo{number}{4} (\bibinfo{year}{2014}),
  \bibinfo{pages}{1030--1081}.
\newblock
\href{https://doi.org/10.1016/j.ijforecast.2014.08.008}{doi:\nolinkurl{10.1016/j.ijforecast.2014.08.008}}


\bibitem[Wong et~al\mbox{.}(2023)]%
        {wong2023deep}
\bibfield{author}{\bibinfo{person}{Annie Wong}, \bibinfo{person}{Thomas
  B{\"a}ck}, \bibinfo{person}{Anna~V Kononova}, {and} \bibinfo{person}{Aske
  Plaat}.} \bibinfo{year}{2023}\natexlab{}.
\newblock \showarticletitle{Deep multiagent reinforcement learning: Challenges
  and directions}.
\newblock \bibinfo{journal}{\emph{Artificial Intelligence Review}}
  \bibinfo{volume}{56}, \bibinfo{number}{6} (\bibinfo{year}{2023}),
  \bibinfo{pages}{5023--5056}.
\newblock
\href{https://doi.org/10.1007/s10462-022-10299-x}{doi:\nolinkurl{10.1007/s10462-022-10299-x}}


\bibitem[Wu et~al\mbox{.}(2016)]%
        {wu2016automated}
\bibfield{author}{\bibinfo{person}{Tony Wu}, \bibinfo{person}{Anthony~D
  Joseph}, {and} \bibinfo{person}{Stuart~J Russell}.}
  \bibinfo{year}{2016}\natexlab{}.
\newblock \emph{\bibinfo{title}{Automated Pricing Agents in the On-Demand
  Economy}}.
\newblock \bibinfo{thesistype}{Master's\ thesis}. \bibinfo{school}{University
  of California, Berkeley}.
\newblock
\urldef\tempurl%
\url{https://www2.eecs.berkeley.edu/Pubs/TechRpts/2016/EECS-2016-57.html}
\showURL{%
\tempurl}


\bibitem[Zhou et~al\mbox{.}(2023)]%
        {zhou2023multi}
\bibfield{author}{\bibinfo{person}{Ziyuan Zhou}, \bibinfo{person}{Guanjun Liu},
  {and} \bibinfo{person}{Ying Tang}.} \bibinfo{year}{2023}\natexlab{}.
\newblock \showarticletitle{Multi-agent reinforcement learning: Methods,
  applications, visionary prospects, and challenges}.
\newblock \bibinfo{journal}{\emph{arXiv preprint arXiv:2305.10091}}
  (\bibinfo{year}{2023}).
\newblock
\href{https://doi.org/10.48550/arXiv.2305.10091}{doi:\nolinkurl{10.48550/arXiv.2305.10091}}


\end{thebibliography}

\newpage
\onecolumn

\appendix
\begin{appendices}

\section{Dataset Description}
\label{sec:apx:dataset_description}

Table~\ref{tab:dataset_overview} provides an overview of the Online Retail II dataset used in this study, including the column names, data types, and brief descriptions. This dataset contains 1,067,371 transactions recorded between December 2009 and December 2011 by a UK-based online retailer. 

\vspace{3mm}
\noindent
\textbf{Table Scope:} While the full dataset includes both B2B and B2C transactions, only B2B transactions (i.e., those with a valid \texttt{Customer ID}) were retained for modeling purposes. All exploratory statistics and preprocessing steps referenced in the methodology are derived from this filtered subset unless stated otherwise.

\begin{table}[H]
\centering
\begin{tabular}{|p{2.5cm}|p{1.8cm}|p{11.5cm}|}
\hline
\textbf{Column Name} & \textbf{Data Type} & \textbf{Description} \\
\hline
\texttt{InvoiceNo} & String & Unique invoice number. Cancellations are indicated by a prefix 'C'. \\
\texttt{StockCode} & String & Unique product code. \\
\texttt{Description} & String & Text description of the product. \\
\texttt{Quantity} & Integer & Number of units sold (can be negative for returns). \\
\texttt{InvoiceDate} & Datetime & Timestamp of the transaction (down to minute precision). \\
\texttt{UnitPrice} & Float & Price per unit of product (in GBP). \\
\texttt{CustomerID} & Integer & Unique identifier for registered customers. Missing for anonymous retail purchases. \\
\texttt{Country} & String & Country of the customer. \\
\hline
\end{tabular}
\caption{Overview of columns in the Online Retail II dataset.}
\label{tab:dataset_overview}
\end{table}

\noindent
A total of 5,243 unique products and 5,942 unique customers were identified in the full dataset. After filtering for valid B2B transactions, 77.2\% of rows remained. Further preprocessing steps, such as the removal of returns (negative quantities), transactions with zero or negative prices, and aggregation at weekly intervals, are described in the main text (see Section~\ref{sec:preprocessing}).

\section{Exploratory Data Analysis Details}
\label{sec:apx:eda}

The following list summarizes key statistics identified in the EDA. All values are based on the dataset filtered for standard products (5-digit stock codes) unless stated otherwise.

\begin{itemize}
    \item 92.1\% of transactions originate from UK customers.
    \item 22.8\% of transactions are from anonymous customers (without Customer ID).
    \item Non-registered customer prices are on average 55.7\% higher; some product categories (e.g., mugs) show 142.3\% higher pricing.
    \item The dataset contains 5,243 unique products and 5,942 registered customers.
    \item 1.7\% of transactions are cancellations, 2.0\% are returns.
    \item Transaction volume spikes during calendar weeks 47–52 (holiday season), with Thursdays and midday showing peak activity.
\end{itemize}

\subsection{Statistical Formulas Used in EDA}

\begin{table}[H]
\centering
\renewcommand{\arraystretch}{1.5}
\small
\begin{tabular}{|p{4.2cm}|p{6cm}|p{6.5cm}|}
\hline
\textbf{Metric} & \textbf{Formula} & \textbf{Result / Interpretation} \\
\hline

\textbf{Volume Discount Calculation} &
\makecell[c]{$\displaystyle \text{Discount}_{\text{bucket}} = \frac{P_{\text{prev}} - P_{\text{curr}}}{P_{\text{prev}}} \times 100\%$\vspace{1.2mm}} & 
Prices dropped by 58.5\% from the 1–10 item bucket to the 10–50 item bucket, confirming strong volume-based pricing incentives. \\

\hline
\textbf{Price Volatility (Coefficient of Variation)} &
\makecell[c]{$\displaystyle CV = \frac{\sigma}{\mu} = \frac{\left( \frac{1}{n} \sum_{t=1}^{n} (V_t - \bar{V})^2 \right)^{1/2}}{\bar{V}}$} &
Average CV is 0.39; 0.2\% of products exceed CV > 2.0, indicating high promotional volatility. \\

\hline
\textbf{Customer Type Price Differential} &
\makecell[c]{$\displaystyle \text{Price Diff}_{\%} = \frac{P_{\text{without ID}} - P_{\text{with ID}}}{P_{\text{with ID}}} \times 100\%$} &
Non-registered customers pay 55.7\% more on average; the MUG category has the largest markup at 142.3\%. \\

\hline
\textbf{Price Ratio (Customer Type)} &
\makecell[c]{$\displaystyle \text{Price Ratio} = \frac{P_{\text{without ID}}}{P_{\text{with ID}}}$} &
Ratios peak at 1.78× during promotional months (e.g., Dec 2011), highlighting potential temporal discrimination. \\

\hline
\textbf{Category Price Differential} &
\makecell[c]{$\displaystyle \text{Markup}_{\text{cat}} = \left( \frac{P_{\text{cat,without ID}}}{P_{\text{cat,with ID}}} - 1 \right) \times 100\%$} &
MUG products show the highest markup (142.3\%), followed by CUSHIONS (69.1\%) and LIGHTS (66.3\%). \\

\hline
\textbf{Quantity-Based Pricing Model} &
\makecell[c]{$\displaystyle P(q) = P_{\text{base}} \cdot f(q) \cdot g(\text{CustomerType})$} &
Model shows $g(\text{Without ID}) \approx 1.56 \cdot g(\text{With ID})$, confirming segmentation by customer type. \\

\hline
\end{tabular}
\caption{Overview of key pricing-related formulas used in EDA with corresponding insights.}
\label{tab:eda_formulas}
\end{table}
\vspace{4mm}

\begin{figure}[H]
  \centering
  \includegraphics[width=0.85\textwidth]{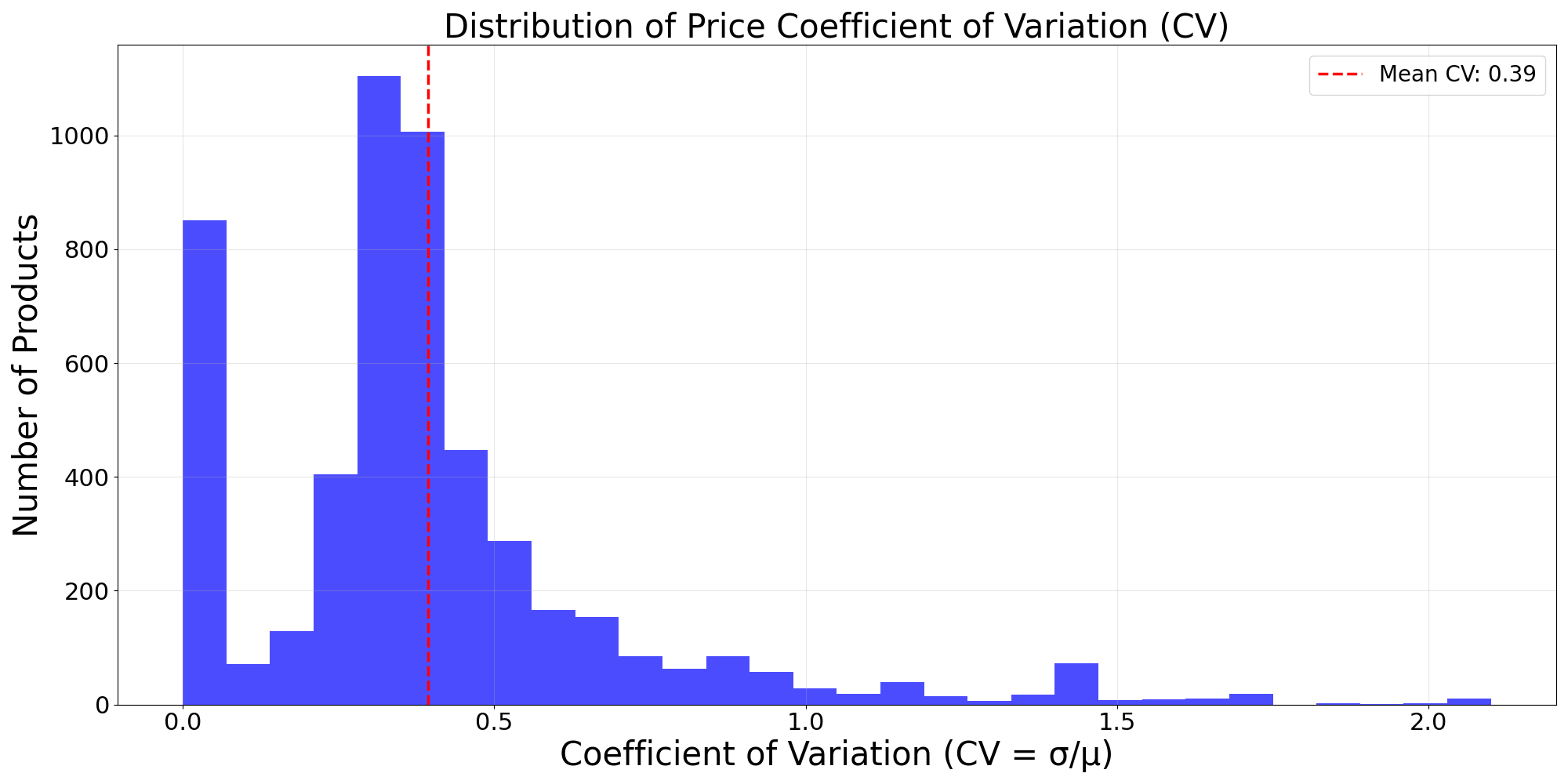}
  \caption{Distribution of price volatility measured by coefficient of variation (CV).}
  \label{fig:price_volatility_distribution_appendix}
\end{figure}

\section{Formulas Used in Preprocessing}
\label{sec:apx:preprocessing_formulas}

The following formulas were used to derive engineered features during preprocessing. These transformations aim to capture temporal trends, volatility, and relative price positioning across product categories.

\begin{table}[H]
\centering
\renewcommand{\arraystretch}{1.6}
\small
\begin{tabular}{|p{3.7cm}|p{4.5cm}|p{8cm}|}
\hline
\textbf{Feature} & \textbf{Formula} & \textbf{Interpretation} \\
\hline

\textbf{Quantity Rolling Mean (QRM)} &
$\displaystyle \text{QRM}_k(t) = \frac{1}{k} \sum_{i=1}^{k} Q_{t - i}$ &
Captures average quantity sold over the previous $k$ weeks, used for lag-based demand modeling. \\

\hline
\textbf{Trend Indicator} &
$\displaystyle \text{Trend}_t = \text{QRM}_4(t) - \text{QRM}_2(t)$ &
Compares medium-term vs short-term average demand to measure directional change. \\

\hline
\textbf{Acceleration} &
$\displaystyle \text{Acceleration}_t = Q_{t-1} - \text{QRM}_2(t)$ &
Measures deviation from short-term average, capturing momentum. \\

\hline
\textbf{Volatility (Rolling Std. Dev.)} &
$\displaystyle \text{Volatility}_t = \sqrt{ \frac{1}{k} \sum_{i=1}^{k} (Q_{t - i} - \bar{Q})^2 }$ &
Estimates fluctuation in recent weekly demand. \\

\hline
\textbf{Price vs Category Average (PVC)} &
$\displaystyle \text{PVC}_{Avg} = \frac{\text{Price}}{\mathbb{E}[\text{Price}_{\text{cluster}}]}$ &
Measures a product's price relative to the average of its semantic cluster (e.g., mugs). \\

\hline
\end{tabular}
\caption{Derived feature formulas used in the preprocessing pipeline.}
\label{tab:preprocessing_formulas}
\end{table}

\noindent
\textbf{Variable Definitions:} $Q_{t-i}$ denotes the quantity sold $i$ weeks before week $t$; $\bar{Q}$ is the mean quantity over the last $k$ weeks; $\mathbb{E}[\text{Price}_{\text{cluster}}]$ refers to the average price of all products within a semantically clustered product group (e.g., mugs, candles); and $\text{Price}$ is the observed price of the product.

\clearpage

\section{Evaluation Metrics Formulas}
\label{sec:apx:evaluation_metrics}

The following table formally defines the evaluation metrics used to assess MARL agent performance in dynamic pricing environments.

\begin{table}[H]
\centering
\renewcommand{\arraystretch}{1.5}
\small
\begin{tabular}{|p{4cm}|p{5cm}|p{8cm}|}
\hline
\textbf{Metric} & \textbf{Formula} & \textbf{Interpretation} \\
\hline

\textbf{Revenue per Agent} &
$\displaystyle R_{agent} = \sum_{t=1}^{T} P_t \cdot Q_t$ &
Total revenue accumulated by an agent over $T$ time steps, where $P_t$ is the price and $Q_t$ is the sold quantity at week $t$. \\


\hline
\textbf{Nash Equilibrium Proximity} &
$\displaystyle \Delta NE = 1 - \min\left(1, \overline{\Delta P} \cdot 10\right)$ &
Measures average absolute price change ratio over recent periods. Closer to 1 implies proximity to equilibrium. \\

\hline
\textbf{Optimality Gap} &
$\displaystyle \text{Gap} = \frac{R_{max} - R_{agent}}{R_{max}}$ &
Relative performance gap between an agent’s revenue and the observed maximum. Lower is better. \\


\hline
\textbf{Gini Coefficient} &
$\displaystyle Gini = \frac{\sum_{i=1}^{N} \sum_{j=1}^{N} |R_i - R_j|}{2N \sum_{i=1}^{N} R_i}$ &
Measures inequality in revenue distribution across agents. A value of 0 indicates perfect equality; values approaching 1 reflect higher inequality. \\

\hline
\textbf{Social Welfare} &
$\displaystyle SW = \sum_{i=1}^{N} R_i \cdot (1 - Gini)$ &
Aggregate market revenue adjusted by fairness (Gini coefficient), rewarding both efficiency and equity. \\

\hline
\textbf{Jain’s Fairness Index} &
$\displaystyle J = \frac{(\sum_{i=1}^{N} R_i)^2}{N \cdot \sum_{i=1}^{N} R_i^2}$ &
Evaluates fairness of revenue distribution across agents, ranging from $1/N$ (worst) to $1$ (best). \\

\hline
\textbf{Market Share Evolution} &
$\displaystyle MS_{i,t} = \frac{R_{i,t}}{\sum_{j=1}^{N} R_{j,t}}$ &
Agent $i$’s share of total market revenue at time $t$. Tracks competitive dynamics. \\

\hline
\textbf{Price Volatility} &
$\displaystyle \text{MeanAbsChange} = \frac{1}{T} \sum_{t=2}^{T} |\frac{P_t - P_{t-1}}{P_{t-1}}|$ &
Mean absolute price change over time, complemented by standard deviation and max change per agent-product pair. \\


\hline
\textbf{Price Convergence} &
$\displaystyle \text{Convergence} = 1 - \frac{\sigma_P}{\max(P)}$ &
Measures how closely agent prices align in the final simulation period. A value close to 1 indicates strong convergence (low price dispersion), while values near 0 indicate significant divergence in pricing strategies. \\

\hline
\textbf{Adjustment Magnitude} &
$\displaystyle \text{AdjMag} = \frac{1}{T{-}1} \sum_{t=2}^{T} \left| \frac{P_t - P_{t-1}}{P_{t-1}} \right|$ &
Mean absolute percentage change in price between consecutive weeks. Captures how intensely agents adjust their prices over time. \\

\hline
\textbf{Adjustment Frequency} &
$\displaystyle \text{AdjFreq} = \frac{1}{T{-}1} \sum_{t=2}^{T} \mathbb{1}\left[ \left| \frac{P_t - P_{t-1}}{P_{t-1}} \right| > \tau \right]$ &
Proportion of weeks with a price change exceeding 1\% ($\tau = 0.01$). Reflects how frequently agents make meaningful pricing updates. \\

\hline
\end{tabular}
\caption{Evaluation metric formulas used for performance assessment of MARL agents.}
\label{tab:evaluation_metrics_formulas}
\end{table}

\noindent
\textbf{Variable Definitions:}
\begin{itemize}
\item $P_t$: Price set by the agent at time step $t$.
\item $Q_t$: Quantity sold by the agent at time step $t$.
\item $R_{i}$: Revenue of agent $i$ over the evaluation period.
\item $R_{i,t}$: Revenue of agent $i$ at time step $t$.
\item $R_{max}$: Maximum observed revenue achieved by any agent in the experiment.
\item $\mu_P$: Mean of the price series over time.
\item $\sigma_P$: Standard deviation of the price series over time.
\item $\overline{\Delta P}$: Average absolute percentage change in price between consecutive time steps.
\item $R_j$: Revenue of agent $j$ over the evaluation period (paired comparison with $R_i$ for Gini calculation).
\item $\Delta R$, $\Delta P$: Changes in revenue and price between consecutive periods or episodes.
\item $N$: Total number of competing agents in the simulation.
\item $\sigma(\Delta P_{products})$: Standard deviation of price changes across an agent’s product portfolio.
\item $\mathbb{1}[\cdot]$: Indicator function that equals 1 if the condition is true and 0 otherwise.
\item $\tau$: Threshold for detecting significant price changes, set to 0.01 (i.e., 1\%).
\end{itemize}

\clearpage

\section{MARL Agent Architecture Diagrams}
The following diagrams provide an exact overview of the operation of MADDPG, MADQN and QMIX agents.

\subsection{MADDPG Architecture}
\label{sec:apx:maddpg_architecture}
\begin{figure*}[htbp]
    \centering
    \includegraphics[width=0.9\textwidth]{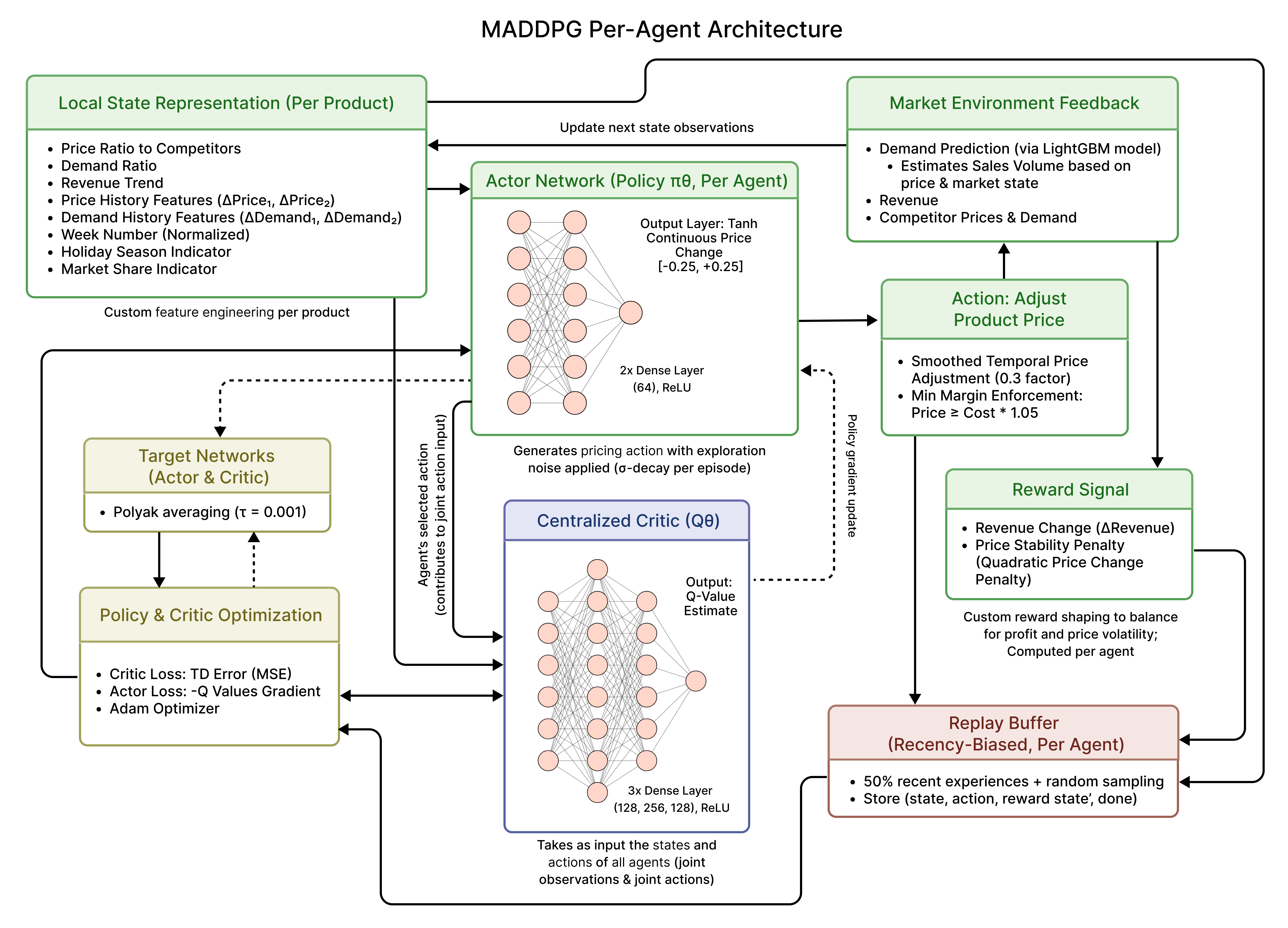}
    \caption{MADDPG Per-Agent Architecture. Each agent selects continuous price changes using a local Actor Network informed by engineered state features (such as price ratios, demand trends, market share). Actions are smoothed and constrained, then evaluated by a centralized Critic receiving joint states and actions. Learning is stabilized via a replay buffer (with recency bias), a TD error-based Critic loss, and Polyak-averaged target networks. The reward function balances revenue gains with a quadratic penalty on price instability, supporting coordinated yet adaptive pricing behaviour.}
    \label{fig:maddpg_architecture}
\end{figure*}

\clearpage

\subsection{MADQN Architecture}
\label{sec:apx:madqn_diagram}
\begin{figure*}[htbp]
    \centering
    \includegraphics[width=\textwidth]{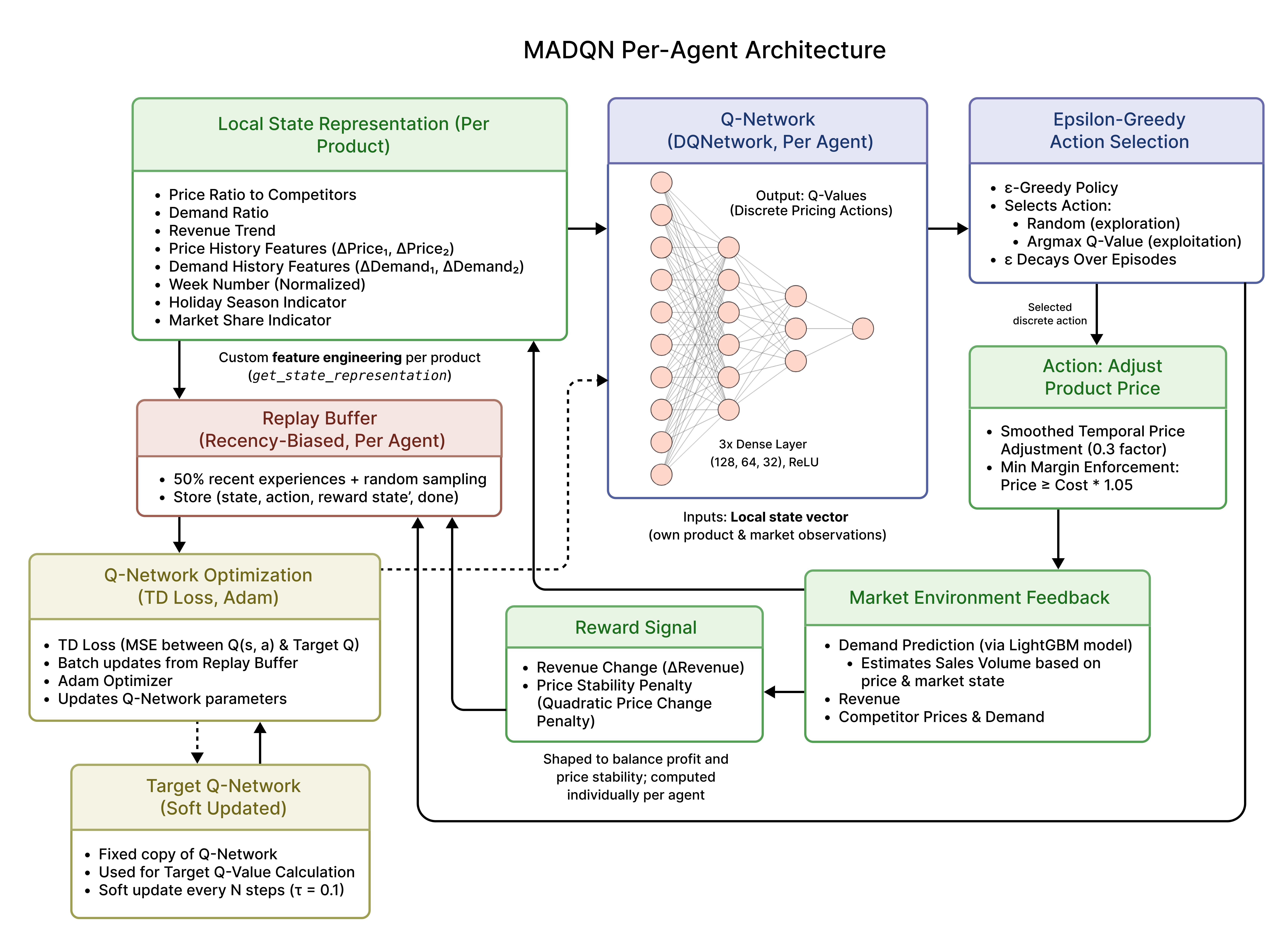}
    \caption{MADQN Per-Agent Architecture. Each agent independently learns discrete pricing actions through a local Q-Network informed by engineered state features. Actions are selected using an $\varepsilon$-greedy policy, then smoothed and constrained to enforce profitability. The agent receives individualized reward signals combining revenue change and a quadratic penalty on price instability. Learning is stabilized via a recency-biased replay buffer, TD loss optimization, and soft target network updates.}
    \label{fig:madqn_architecture}
\end{figure*}

\clearpage

\subsection{QMIX Architecture}
\label{sec:apx:qmix_diagram}
\begin{figure*}[htbp]
    \centering
    \includegraphics[width=\textwidth]{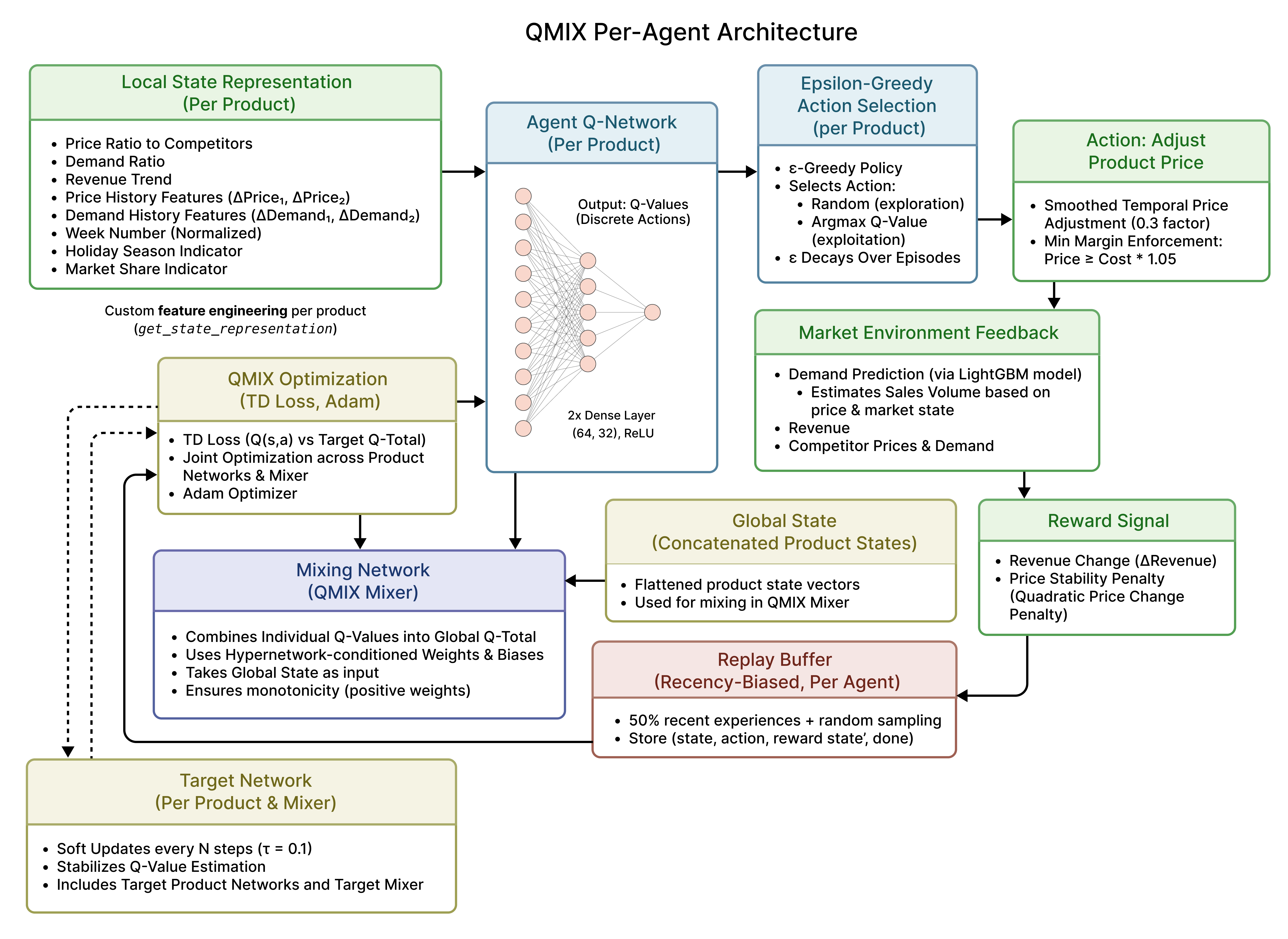}
    \caption{QMIX Per-Agent Architecture. Each pricing agent uses a local Q-Network to learn discrete price adjustments based on engineered state features. Actions are selected via $\varepsilon$-greedy exploration and post-processed through domain-specific constraints. A centralized Mixing Network then combines individual Q-values into a joint Q-total, using hypernetwork-conditioned weights based on the concatenated global state. The network enforces monotonicity to ensure consistency between decentralized decisions and centralized optimization. Training relies on a recency-biased replay buffer, joint TD loss minimization, and soft-updated target networks for both agent and mixer components. Learning is driven by a shared global reward signal balancing revenue changes and price stability, enabling coordinated but decentralized pricing strategies.}
    \label{fig:qmix_architecture}
\end{figure*}

\end{appendices}

\end{document}